\documentclass{article}

\usepackage{arxiv}

\usepackage[utf8]{inputenc} 
\usepackage[T1]{fontenc}    
\usepackage{hyperref}       
\usepackage{url}            
\usepackage{booktabs}       
\usepackage{amsfonts}       
\usepackage{nicefrac}       
\usepackage{microtype}      
\usepackage{lipsum}
\usepackage{graphicx}
\graphicspath{ {./images/} }
\usepackage{tabularx}
\usepackage{array}

\title{A Study on Real-time Object Detection using Deep Learning}

\author{
  Ankita Bose \\
  Department of Computer Science and Engineering\\
  GITAM University\\
  India, 561203 \\
  \texttt{abose@gitam.edu} \\
   \And
  Jayasravani Bhumireddy \\
  Department of Computer Science and Engineering\\
  GITAM University\\
  India, 561203 \\
  \texttt{jbhumire2@gitam.in} \\
  \And
  Naveen N \\
  Department of Computer Science and Engineering\\
  GITAM University\\
 India, 561203 \\
  \texttt{nnarasim2@gitam.in} \\
}

\begin{document}
\maketitle
\begin{abstract}
Object detection has compelling applications over a range of domains, including human-computer interfaces, security and video surveillance, navigation and road traffic monitoring, transportation systems, industrial automation healthcare, the world of Augmented Reality (AR) and Virtual Reality (VR), environment monitoring and activity identification. Applications of real time object detection in all these areas provide dynamic analysis of the visual information that helps in immediate decision making. Furthermore, advanced deep learning algorithms leverage the progress in the field of object detection providing more accurate and efficient solutions. There are some outstanding deep learning algorithms for object detection which includes, Faster R CNN(Region-based Convolutional Neural Network),Mask R-CNN, Cascade R-CNN, YOLO (You Only Look Once), SSD (Single Shot Multibox Detector), RetinaNet etc. This article goes into great detail on how deep learning algorithms are used to enhance real time object recognition. It provides information on the different object detection models available, open benchmark datasets, and studies on the use of object detection models in a range of applications. Additionally, controlled studies are provided to compare various strategies and produce some illuminating findings. Last but not least, a number of encouraging challenges and approaches are offered as suggestions for further investigation in both relevant deep learning approaches and object recognition.
\end{abstract}


\section{Introduction}
In recent years, significant progress has been made in the field of computer vision, thanks to the advancements in machine learning techniques. Machine learning provides a robust solution to address various computer vision problems. Specifically, in the context of object detection. The objective is to accurately locate and identify instances of specific classes within digital images or videos [1]. The objects must be categorised and localised in order to be recognised. Object Detection involves identifying and locating objects in images or videos. There are some major aspects that we should focus on while designing an object detection algorithm. For instance, a well annotated dataset with high quality and diverse data will be helpful to any object detection model to ensure higher accuracy of prediction. Again bounding box [2] plays an important role in tracking an object in images and video frames along with object's possible extent. Therefore knowing the exact spatial location of the object is very important. In this regard, anchor boxes [3] are really important to predefine the size and shape of bounding box. Bounding boxes have many applications, which includes surveillance system, object counting, autonomous vehicles, crowd analysis, robotics etc. Furthermore, we can consider the model architecture as the backbone of the object detection algorithm. It directly impacts the speed and accuracy and hence the overall performance of that particular algorithm. In this study we have added a comparative analysis of the model architectures of different object detection algorithms. Another important aspect is backbone network [4] of any real-time object detection model. Backbone network is responsible to extract features from images, which is very essential for any real-time object detection algorithm. Therefore, backbone network has to be light weight and computationally efficient to meet accuracy along with speed requirement for real-time detection. In addition, specialized hardware that supports parallel processing can significantly improve the performance of the object detection model.

The historical account of deep learning with its advancement in object detection notably real-time object detection would help us to understand the way where we are today. It all started with the conventional Artificial Neural Network in 1940s [5]. The purpose was to mimic the behaviour and functioning of human brain. Though it was introduced in 1940 but it had experienced a boost in 1980s-1990s [6] with the introduction of backpropagation. Unfortunately backpropagation doesn't work well with large number of hidden layers. The presence of non-convex error surface with local optima, flat spots, highly multi-diamensional data make it challenging to train a deep neural network (DNN). Moreover DNN suffers the unstable gradient problem when different layers tend to learn at different speeds. Backpropagation works on the local gradient calculation. Therefore, when network depth increases, backpropagation is not a good choice for training that network. On the other hand, in early 2010s object detection was mostly a single window approach based on conventional computer vision techniques. Methods like parametric flow models [7], intra-frame matching [8], histogram of oriented gradients [9] were very popular. However, human visual information processing capability has a layered hierarchial structure to process human visual perception. Deep architecture of DNN is capable of nonlinear processing. Hence, it has been applied successfully on various applications of object detection since 2012. In 2012, A Krizhevsky et. al.[10] first introduced AlexNet. They trained a deep convolutional neural network on ImageNet data set that spans over 1000 classes and applied it to classify 1.3 million high resolution images. In 2014, Ross Girshick et al. proposed a Region based convolutional neural network (RCNN) [11] and applied it on  PASCAL VOC dataset. RCNN works on three stages, First region proposals (2000 proposals) are identified for an input image, second, from each region features are extracted after passing it through CNN and finally region specific features are classified to identify the object class. To deal with the speed issues of RCNN, Ross Girshick et al. proposed Fast-RCNN in 2015 [12]. In Fast-RCNN, instead of considering 2000 region proposals for every image, a convolutional feature map is generated using CNN. From the feature map region proposals are identified followed by feature extraction and classification. In the same year (2015), Shaoqing Ren et al. introduced Faster-RCNN [13]. In Faster RCNN, they applied Region Proposal Network (RPN) to predict the region proposals. Faster-RCNN achieved improved performance in terms of speed and accuracy over Fast-RCNN. In 2016[14], Wei Liu et al. introduced SSD (Single Shot Multibox Detector). Which in turn using a single deep learning neural network that combines the prediction of multiple bounding boxes of different scale and aspect ratio from various feature maps. Some other lightweight modified SSD architectures are also available. In this literature we have discussed them with greater detail. During the same year 2016 [15], Joseph Redmon et al. introduced YOLO (You only look once). In YOLO, an input image is divided into a grid followed by the prediction of bounding boxes and class probabilities done in a single pass.  Since 2016, many variations of YOLO (e.g V2,V3, V4, V5, V6, V7) has been introduced. In this literature we have provided a detail discussion on the evolution of YOLO in 2016-2024.  For the efficient use of deep neural network architectures in mobile and embedded systems, Andrew G. Howard et al. in 2017 [16] introduced MobileNet. In which, depthwise separable convolution operations significantly improve the accuracy and speed of the model. Apart from this, some light-weight architecture based advanced model has been introduced, such as RetinaNet in 2017 [17], CenterNet in 2019 [18], EfficientDet in 2020 [19] etc.

The subsequent sections of this article are organised as follows. Section II provides a brief architectural overview of the deep learning models for real time object detection, in Section III we explore the real-time applications of deep learning models focusing both generic and salient object detection, Section IV showcase the analysis of the existing works and scope for future research, Section V summarizes the main findings.

\section{Architectural Overview of The Deep Learning Models}
\subsection{Convolutional Neural Network}
In 1998 [20], LeCun, Y et al designed LeNet-5, a convolutional neural network for handwritten digit recognition. This neural network model is trained on gradient based methods along with back propagation. A Convolutional Neural Network (CNN or ConvNet) is a class of neural networks that is typically used for challenging image driven pattern recognition tasks.

\begin{itemize}
  \item \textbf{Architecture}
  A typical Convolutional Neural Network (CNN) consists of four primary layers: the convolution layer, pooling layer, fully connected layer, and nonlinear layer. The convolution layer applies a kernel filter to the input image to compute convolutions and extract fundamental features like edges and textures. The pooling layer combines outputs from successive convolution layers, reducing spatial dimensions and computational complexity. The fully connected layer, also known as the convolutional output layer, synthesizes features from earlier layers to produce predictions. Lastly, the nonlinear layer employs activation functions to determine the network's output, enabling tasks such as binary classification (e.g., yes or no).  Common activation functions used in CNNs include Sigmoid, Tanh, ReLU (Rectified Linear Unit), and their variants such as Leaky ReLU, Noisy ReLU, and Parametric ReLU. These functions introduce nonlinearity, enabling CNNs to learn and represent complex patterns effectively. The design and architecture of CNNs are heavily inspired by the organization and functioning of the human visual cortex, mimicking neural connections in the brain to process visual information.

  In Figure 1, (b) illustrates the Convolution Operation:\\
    \textbf{Input Matrix (4x4):} Represents the input data or an intermediate feature map during processing.\\
    \textbf{Kernel (3x3):} A small filter matrix that slides over the input, performing element-wise multiplication to generate the output matrix (2x2).\\
    \textbf{Stride:} Defines the step size of the kernel as it moves across the input. For instance, a stride of 1 means the kernel shifts one step at a time.\\
    \textbf{Padding:} Involves adding extra rows or columns to the input matrix to control the dimensions of the resulting feature map.
\end{itemize}

\begin{figure}[ht]
  \centering
  \includegraphics[width=0.8\textwidth, height=3in]{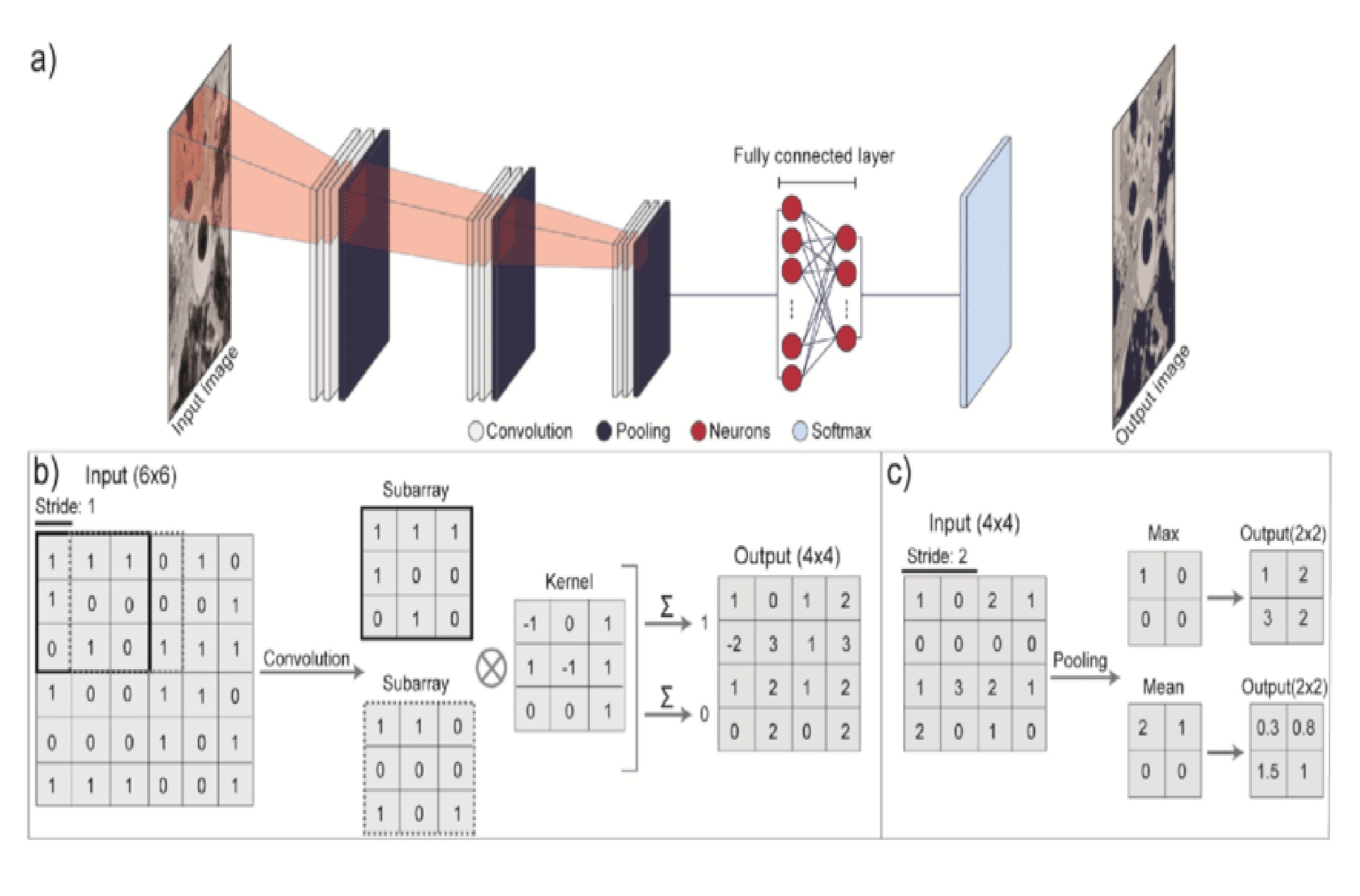}
  \caption{a) Architecture of CNN, b)Illustrates Convolution Operation, c) Illustrates pooling operations}
  \end{figure}

\begin{table}[ht]
  \scriptsize
  \centering
  \begin{tabularx}{\textwidth}{|X|X|X|}
  \hline
  \textbf{CNN Variation} & \textbf{Details} & \textbf{Application} \\ \hline
  \textbf{LeNet} (1998) & First CNN architecture for digit recognition. & Recognition of digits (for instance, MNIST dataset)[41]. \\ \hline
  \textbf{AlexNet} (2012) & Advanced the use of deep CNNs featuring ReLU activation and training on GPUs & Visual categorization (e.g., ImageNet contest)[42][43] \\ \hline
  \textbf{ZFNet} (2013) & Enhanced AlexNet by modifying filter dimensions and integrating feature visualization techniques. & Image classification and interpretability [44] \\ \hline
  \textbf{VGGNet} (2014) & Used deeper architectures with small 3x3 filters. Demonstrated the importance of network depth. & Image categorization, attribute extraction for transfer learning.[45] \\ \hline
  \textbf{GoogLeNet} (Inception-v1) (2014) & Introduced inception modules with multi-scale convolutions and dimensionality reduction (1x1 convolutions). & Object detection and image categorization [46]. \\ \hline
  \textbf{ResNet} (2015) & Implemented residual learning and skip connections to support extremely deep networks (reaching 152 layers). & Image categorization, object recognition, healthcare imaging [47].\\ \hline
  \textbf{Fully Convolutional Networks} (2015) & Converted fully connected layers into convolutional layers for image segmentation. & Semantic segmentation, analysis of medical images [48].\\ \hline
  \textbf{Inception-v3} (2015) & Enhanced Inception units featuring factorized convolutions. & Visual recognition, object identification.\\ \hline
  \textbf{U-Net} (2015) & Crafted for segmentation utilizing a U-shaped encoder-decoder structure. & Segmentation of medical images, satellite imagery analysis [49].\\ \hline
  \textbf{MobileNet} (2017) & Introduced depthwise separable convolutions for lightweight models suitable for mobile devices. & Applications for mobile and edge devices (e.g., facial recognition)[50].\\ \hline
  \textbf{SENet (Squeeze-and-Excitation Network)} (2017) & Introduced channel-wise attention mechanisms to enhance feature selection. & Object detection, image categorization [51].\\ \hline
  \textbf{DenseNet} (2017) & Implemented dense connectivity that allows layers to share features with every following layer. & Medical imaging, tasks with extensive features [52].\\ \hline
  \textbf{DeepLab} (2018) & Used dilated convolutions and atrous spatial pyramid pooling for semantic segmentation. & Semantic segmentation (e.g., self-driving cars, city development)[53].\\ \hline
  \textbf{ShuffleNet} (2018) & Employed channel shuffling along with grouped convolutions for improved efficiency. & Optimized neural networks for mobile devices [54].\\ \hline
  \textbf{CapsNet} (Capsule Networks) (2018) & Encoded spatial hierarchies with capsules to handle object pose variations. & Image classification, object recognition, and tasks requiring viewpoint invariance [55].\\ \hline
  \textbf{EfficientNet} (2019) & Used neural architecture search (NAS) to scale models efficiently. & Image classification, object recognition, medical imaging [56].\\ \hline
  \textbf{ConvNeXt} (2021) & Improved traditional CNNs with modern techniques like LayerNorm and GELU activation for competitive performance. & image classification and computer vision tasks [57]\\ \hline
  \end{tabularx}
  \caption{Variations of CNN invented so far}
  \end{table}

\subsection{R-CNN}
One of the most renowned object detection models in the category of two-stage detectors is known as R-CNN, which stands for region-based convolutional neural network. R-CNN was introduced by Girshick et al. [21]. RCNN introduced an innovative approach by combining Convolutional Neural Networks (CNNs) with region proposal techniques to effectively identify and localize objects in images. The algorithm begins by generating approximately 2000 region candidates using Selective Search, identifying potential areas where objects might exist. These regions are resized to a fixed dimension and processed by a pre-trained CNN (e.g., AlexNet or VGG) to extract feature vectors. These features are then used by a Support Vector Machine (SVM) for classification[22], while a regressor refines the bounding box coordinates, ensuring accurate object detection and localization. RCNN's capability to both identify and precisely locate objects has made it applicable in diverse domains such as autonomous vehicles, healthcare, security, retail, and automation. However, its limitations include the need to process the CNN multiple times for each region proposal, resulting in slow inference, high storage requirements for deeper layers, and a multi-stage training process, which adds complexity and computational cost. Despite its contributions, the model has notable limitations, such as the need to repeatedly feed the CNN for each region proposal, resulting in slow processing, high storage demands for deeper layers, and inefficiencies in multi-stage training. Upgraded versions like Fast R-CNN, Faster R-CNN, and Mask R-CNN address these challenges by introducing innovations such as RoI pooling and Region Proposal Networks (RPNs) to enhance performance. R-CNN played a pivotal role in bridging the gap between image classification and object localization, paving the way for real-time systems.

\begin{itemize}
  \item \textbf{Architecture}
  The architecture of R-CNN integrates region proposal methods with deep learning-based feature extraction and classification in a structured manner. The process begins with generating approximately 2000 region proposals using techniques like Selective Search, identifying potential areas where objects may be present in the image. To standardize the input size for the model, these regions (Regions of Interest, or RoIs) are resized to 224x224 pixels. These resized regions are then passed through a pre-trained convolutional neural network, such as AlexNet or VGG, which transforms the raw image data into rich feature representations, producing high-dimensional feature vectors. Based on these features, an SVM classifier predicts the object category for each region, while a bounding box regressor fine-tunes the coordinates to accurately position the bounding boxes. The final output of the R-CNN model includes a set of bounding boxes, class predictions, and confidence scores for the detected objects. Although this architecture was groundbreaking and highly accurate, it is computationally expensive, as each region proposal must pass through the CNN individually, leading to longer inference times compared to more modern object detection models. The figure below illustrates the RCNN architecture.
\end{itemize}

  \begin{figure}[ht]
  \centering
  \includegraphics[width=0.9\textwidth, height=3in]{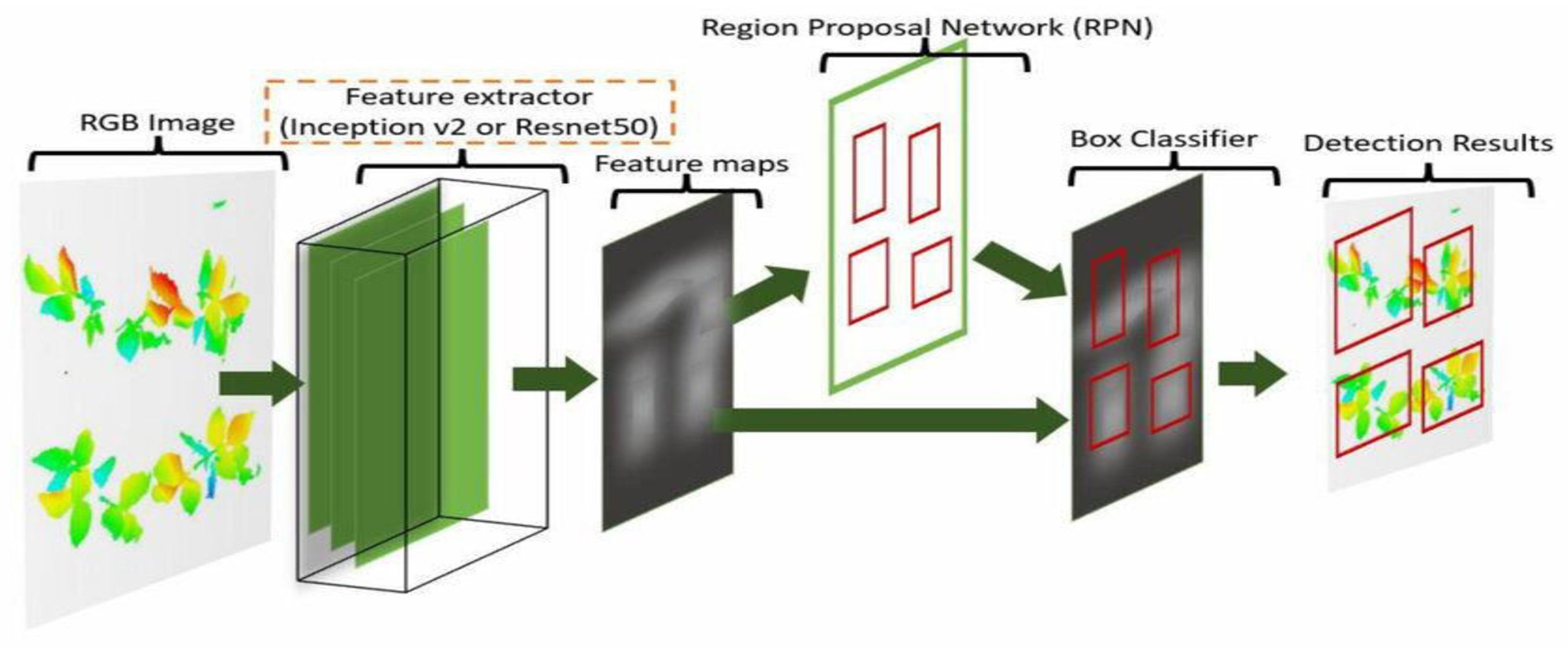}
  \caption{ Architecture of R-CNN}
  \end{figure}

\subsection{Fast R-CNN}
A year later, in 2015, Girshick [12] introduced a faster variant of R-CNN called Fast R-CNN. Fast R-CNN is an advanced object detection algorithm that addresses the limitations of its predecessor, R-CNN. It introduced an innovative framework by integrating classification and localization tasks into a unified structure during training. Unlike R-CNN, Fast R-CNN reduces computational costs by performing the convolutional operation only once for the entire image, rather than for each region proposal individually. The resulting feature map is then shared across all region proposals, significantly improving efficiency. A significant advancement in this field is the introduction of the Region of Interest (RoI) pooling layer, which enables the extraction of uniform-sized feature maps from regions of varying sizes. By combining shared feature extraction, a softmax classifier, and bounding box regression into a single unified framework, Fast R-CNN offers an excellent balance between speed and accuracy. These enhancements make it highly suitable for real-world applications such as autonomous vehicles, surveillance systems, medical imaging, and retail analytics.

\begin{itemize}

  \item \textbf{Architecture}
  Fast R-CNN's architecture is designed to optimize object detection through a streamlined and unified process. It begins with an input layer that accepts an image of any size and processes it using a pre-trained Convolutional Neural Network (CNN), such as VGG16 or ResNet. The CNN produces feature maps that capture the spatial and semantic details of the image. Unlike R-CNN, which processes each region proposal independently, Fast R-CNN generates feature maps only once and reuses them for all region proposals, significantly improving speed and efficiency.\\
  Next, region proposals are generated using methods like Selective Search or Region Proposal Networks (RPNs) to identify potential object locations. These proposals are then processed through the Region of Interest (RoI) pooling layer, a critical component that extracts fixed-size feature maps from each region proposal. By dividing proposals into grids and applying max-pooling, this layer ensures consistent input dimensions for the fully connected layers. The fixed-size feature maps are passed through fully connected layers, which process the data and generate two outputs: a softmax classifier to predict object classes and a bounding box regressor for precise localization. Combined with techniques like non-maximum suppression, these outputs yield the final results, including identified objects, their categories, and refined bounding box coordinates.\\
  This architecture excels due to its efficiency, achieved through shared computations, and its accuracy, enabled by end-to-end training. Fast R-CNN strikes an effective balance between computational speed and detection accuracy, making it a cornerstone of modern object detection systems.

\end{itemize}

  \begin{figure}[ht]
  \centering
  \includegraphics[width=0.9\textwidth, height=2.5in]{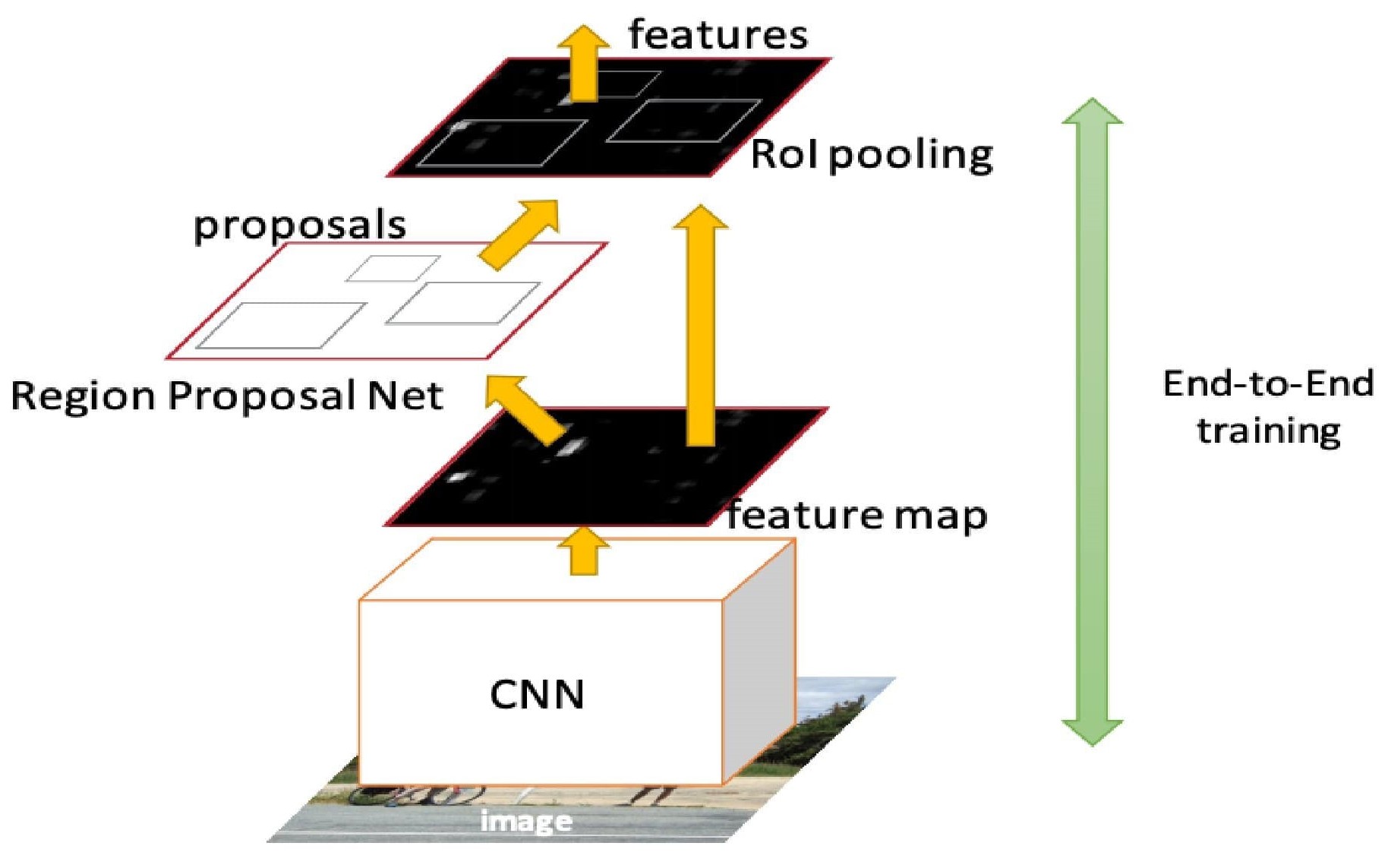}
  \caption{ Architecture of Fast R-CNN}
  \end{figure}

\subsection{Faster R-CNN}
Faster R-CNN [23] is a sophisticated object detection framework that integrates a Region Proposal Network (RPN) with the Fast R-CNN detection algorithm. This unified approach streamlines feature extraction, region proposal generation, and object detection into a single workflow, enhancing both efficiency and accuracy. The process starts by passing an input image through a pre-trained Convolutional Neural Network (CNN), such as ResNet or VGG, to produce feature maps. These feature maps encapsulate essential spatial and semantic information required for object identification. This design enhances the performance of earlier models by integrating region proposal generation directly within the network, significantly lowering computational costs. By uniting feature extraction with end-to-end training, it boosts detection accuracy while maintaining efficient processing. This makes Faster R-CNN well-suited for applications requiring both high precision and scalability.

\begin{itemize}
  \item \textbf{Architecture}
  The Faster R-CNN architecture is a powerful two-stage object detection framework designed for accuracy and efficiency. It begins with a shared convolutional backbone (e.g., ResNet or VGG) to extract feature maps from the input image. These feature maps are processed by the Region Proposal Network (RPN), which generates region proposals using anchor boxes of different sizes and aspect ratios. The RPN assigns an objectness score to each anchor and refines its coordinates through bounding box regression, eliminating the need for external region proposal methods like Selective Search.\\
  In the second stage, the proposed regions are passed through a RoI pooling layer, which extracts fixed-size feature maps from the variable-sized proposals. These feature maps are then processed by fully connected layers to perform object classification (or background detection) and further refine the bounding box coordinates.\\
  The architecture is end-to-end trainable, leveraging shared features between the RPN and the final detection head to improve computational efficiency. Its flexibility in handling objects of varying scales and aspect ratios, combined with its streamlined design, makes Faster R-CNN a leading choice for high-accuracy object detection tasks in diverse applications such as autonomous driving, medical imaging, and video surveillance.
\end{itemize} 

  \begin{figure}[ht]
  \centering
  \includegraphics[width=0.9\textwidth, height=3in]{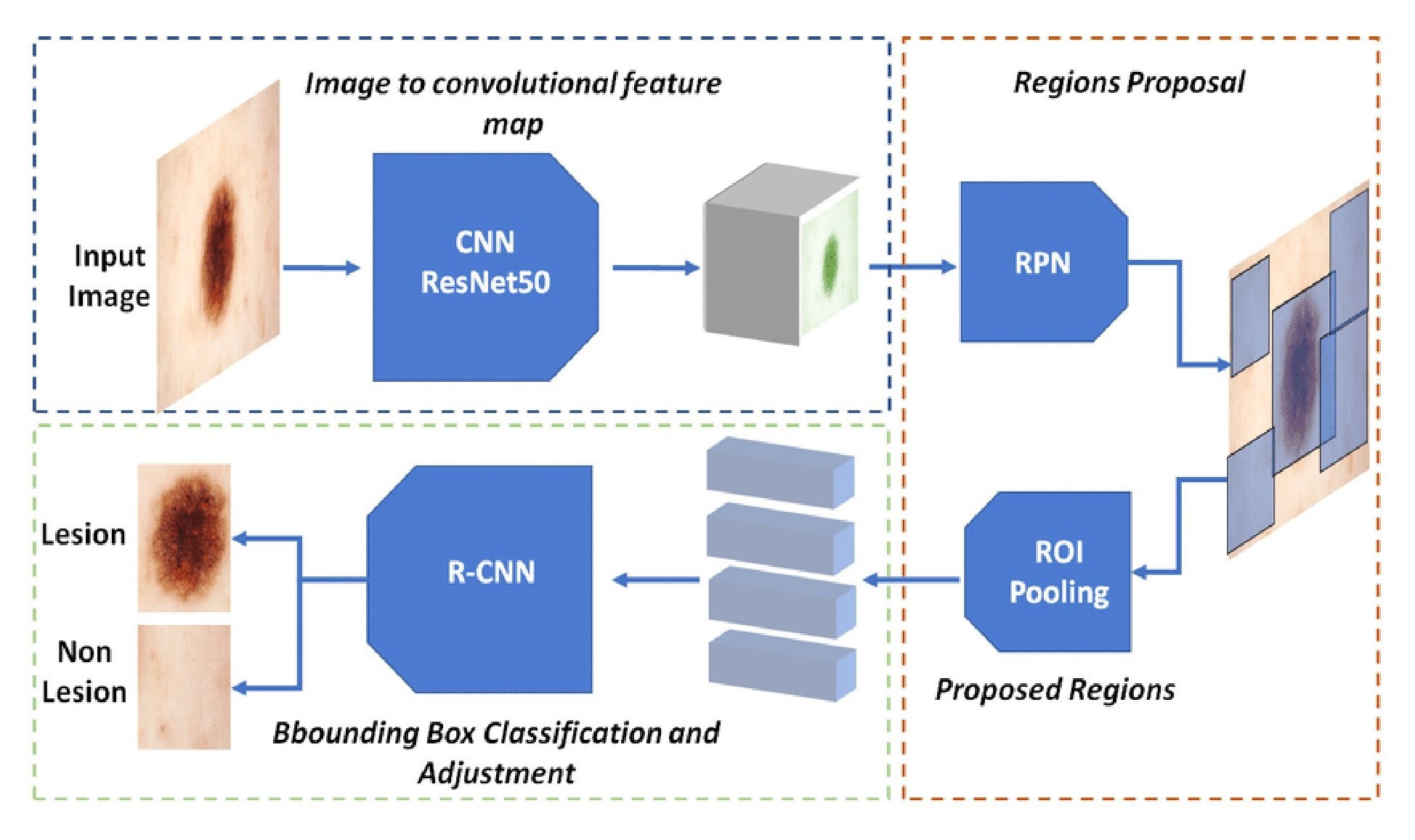}
  \caption{ Architecture of Faster R-CNN}
  \end{figure}

\subsection{YOLO(You Only Look Once)}
YOLO (You Only Look Once) is a widely recognized and efficient object detection method introduced by Joseph Redmon et. al. in 2016 [24]. It revolutionized object detection by framing it as a single regression problem, allowing the model to predict class labels and bounding box coordinates for all objects in an image simultaneously. This streamlined approach makes YOLO significantly faster than earlier methods like R-CNN and Faster R-CNN, which rely on region proposals and multiple processing stages.\\
The primary advantage of YOLO lies in its ability to perform real-time object detection by processing the entire image in a single pass, rather than analyzing smaller segments. It divides the image into a grid, where each cell predicts a bounding box and a confidence score for the objects it contains. These predictions are made simultaneously, greatly increasing the detection speed while maintaining the accuracy.\\
Over the years, YOLO has evolved through multiple versions, from YOLOv1 to the latest YOLOv10, with each iteration improving speed, accuracy, and versatility for a wide range of applications. Its efficiency and ability to detect multiple objects in a single frame have made YOLO one of the most widely used object detection models in real-time applications such as autonomous driving, robotics, video surveillance, and augmented reality.

\begin{itemize}
  \item \textbf{Architecture}
  The YOLO (You Only Look Once) architecture is designed for fast and efficient object detection, allowing the model to predict class labels and bounding box coordinates in a single forward pass. The input image is divided into a grid of cells, with each cell responsible for detecting objects whose centers fall within it. A convolutional neural network (CNN), such as Darknet-53, serves as the backbone, extracting image features that are processed by the detection head. The detection head predicts multiple bounding boxes per grid cell, outputting the box coordinates (x, y, width, height), a confidence score, and class probabilities.\\
  The predictions are combined into a tensor, and post-processing techniques like Non-Maximum Suppression (NMS)[25][26] are applied to remove duplicate or overlapping boxes, retaining only the most accurate ones. YOLO’s architecture streamlines object detection into a unified process, offering significant speed advantages over multi-stage models like those using region proposal networks.\\
  This efficiency makes YOLO well-suited for real-time applications. Additionally, with continuous improvements across its versions (YOLOv1 to YOLOv10), YOLO has enhanced its accuracy and ability to handle increasingly complex detection tasks, cementing its versatility and scalability.  
\end{itemize}

  \begin{figure}[ht]
  \centering
  \includegraphics[width=0.9\textwidth, height=4in]{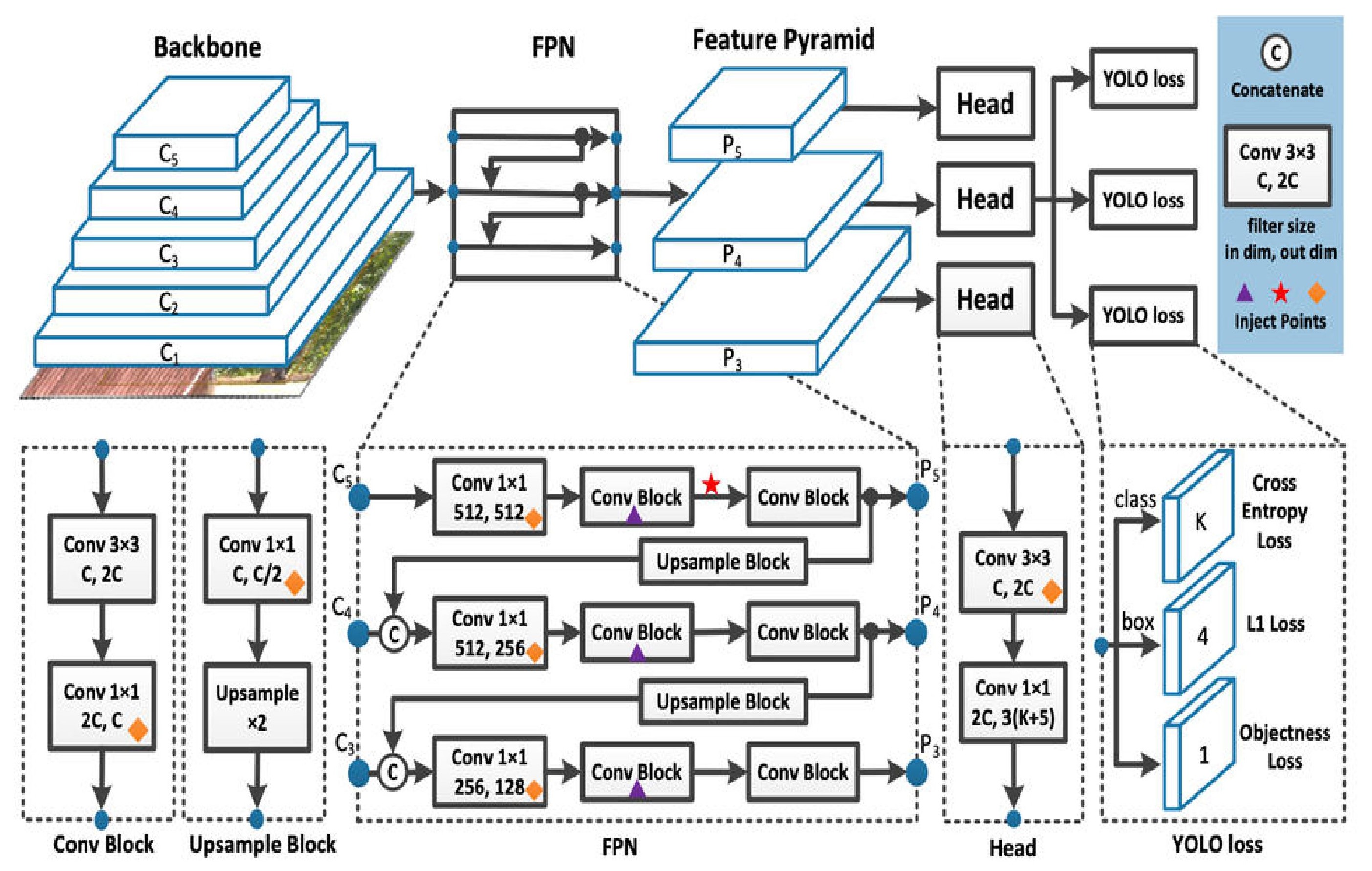}
  \caption{ Architecture of YOLO}
  \end{figure}
  
\clearpage

\begin{table}[ht]
  \scriptsize
  \centering
  \begin{tabularx}{\textwidth}{|X|X|}
  \hline
  \textbf{YOLO Variations} & \textbf{Details} \\ \hline
  \textbf{YOLOv1} (2016) & The first version of YOLO [24] introduced the concept of performing object detection in a single pass through the network. It divided the image into a grid, predicting bounding boxes and class probabilities for each cell. While it was significantly faster than previous methods, YOLOv1 had lower accuracy, especially when detecting smaller objects.\\ \hline
  \textbf{YOLOv2} (2017) & YOLOv2 [27], also known as "YOLO9000," introduced several improvements, including a more powerful backbone network (Darknet-19), higher grid resolution, and the ability to detect over 9,000 object categories. It also incorporated batch normalization and anchor boxes, significantly boosting both detection accuracy and speed. \\ \hline
  \textbf{YOLOv3} (2018) & YOLOv3 [28] improved its architecture by adopting Darknet-53 as the backbone, providing a deeper and more efficient network compared to Darknet-19. It introduced multi-scale predictions, enabling better detection of objects across different sizes. This enhancement significantly improved the model’s ability to identify small objects and increased accuracy, particularly in complex scenarios.\\ \hline
  \textbf{YOLOv4} (2020) & YOLOv3 enhanced its architecture by integrating Darknet-53 as the backbone, offering a deeper and more efficient network than Darknet-19. It introduced multi-scale predictions, allowing for improved detection of objects of varying sizes [29]. These advancements significantly boosted the model with enhanced accuracy, especially in challenging and complex environments.\\ \hline
  \textbf{Variations of YOLOv4} (2020) & \textbf{YOLOv4-CSP}[30]- Uses the CSPDarknet53 backbone for improved feature extraction, optimizing speed and accuracy in real-time applications. \textbf{YOLOv4-tiny}[31]- It is a fast, lightweight version for real-time detection on resource-limited devices, prioritizing speed over accuracy for low-latency applications. \textbf{Scaled-YOLOv4}[32]-Scaled-YOLOv4 optimizes YOLOv4 for scalability, using CSPNet, PANet, and SPP to enhance accuracy and speed. \\ \hline
  \textbf{YOLOv5} (2020) & YOLOv5 [33], developed by Ultralytics, is a widely adopted variant known for its speed, versatility, and ease of use. It introduced multiple model sizes (small, medium, large) to suit different hardware configurations. With an optimized training process, YOLOv5 enhances performance across various datasets, making it highly efficient for real-time object detection applications.\\ \hline
  \textbf{YOLOv6} (2022) & YOLOv6 [34] is a fast, accurate object detection model with anchor-free detection, an efficient backbone, and improved post-processing. Optimized for real-time and industrial applications, it enhances training and inference efficiency, making it ideal for surveillance, autonomous driving, and robotics. Its design ensures high precision and speed for production use.\\ \hline
  \textbf{YOLOv7} (2022) & YOLOv7[35] is a high-speed, accurate object detection model. It features extended efficient layer aggregation networks (E-ELAN) and model scaling strategies for enhanced performance. Designed for real-time applications, it surpasses previous versions in accuracy and inference speed, making it ideal for surveillance, robotics, and autonomous driving, where precision and efficiency are crucial.\\ \hline
  \textbf{YOLOv8} (2023) & YOLOv8 [36], currently in development, aims to enhance efficiency, scalability, and precision for real-time applications. It features an optimized architecture using CSPDarknet and PAN-FPN for improved feature extraction. The model supports object detection, classification, and segmentation in a single framework. Additionally, it ensures compatibility with ONNX, TensorRT, and various edge devices, making it versatile for multi-domain object detection tasks.\\ \hline
  \textbf{YOLOv9} (2024) & Recent research by Wang et al. (2024) on YOLOv9 [37] introduces two key innovations: Programmable Gradient Information (PGI) for preserving gradient data and Generalized Efficient Layer Aggregation Network (GELAN) for efficient feature extraction. These advancements improve accuracy and computational efficiency, making YOLOv9 a cutting-edge solution for diverse applications.\\ \hline
  \textbf{YOLOv10} (2024) & The recent YOLOv10 [38] model introduced by Wang et al., YOLOv10 eliminates Non-Maximum Suppression (NMS) by using dual assignments, enabling faster inference. Its optimized architecture enhances feature extraction, multiscale fusion, and introduces a one-to-many training head and a one-to-one inference head. YOLOv10-S improves speed by 1.8× over RT-DETR-R18, while YOLOv10-B reduces latency by 46\% compared to YOLOv9-C. These advancements make YOLOv10 a cutting-edge solution for real-time object detection.\\ \hline
  \textbf{PP-YOLO} (2020-2021) & PP-YOLO (2020) [39] by PaddlePaddle is an improved YOLOv4 with ResNet50-vd, IoU-aware loss, EMA, incorporates DropBlock for better regularization, and  Uses Matrix NMS, achieving 45.2\% mAP at 72.9 FPS. It is Faster but heavier than YOLOv3, later versions like PP-YOLOv2 (2021)[40] and PP-YOLOE (2022) [41] further improved accuracy and efficiency.\\ \hline
  \end{tabularx}
  \caption{Variations of YOLO invented so far}
  \end{table}

\clearpage

\begin{table}[ht]
  \scriptsize
  \centering
  \begin{tabularx}{\textwidth}{|X|X|X|X|X|X|}
  \hline
  \textbf{Name of the Model(Year)} & \textbf{Base Architecture} & \textbf{Data Set} &\textbf{Model Trained on} &\textbf{Mean-Average Precision (mAp)} &\textbf{Frames Per Second (FPS)}\\ \hline
  \textbf{YOLOv1} (2016) & Inspired by GoogleNet. It has replaced the inception module with a (1 x 1) convolution followed by (3 x 3) convolution filters &  VOC Pascal Dataset. &  Darknet framework & 63.4\% & 45\\ \hline
  \textbf{YOLOv2} (2017) & Darknet-19 backbone of YOLOv2 was inspired by VGG \& ResNet. A typical darknet-19 framework consists of 19 convolutional layers and 5 max-pooling layers & PASCAL VOC 2007 Test Set, COCO & ImageNet, COCO & 67.8\% & 67\\ \hline
  \textbf{YOLOv3} (2018) & Darknet-53 is the backbone of YOLOv3. It has 53 convolution layers with (3 x 3) and (1 x1) convolutional filters and it uses residual connections like ResNet & COCO & COCO, PASCAL VOC 2007 and VOC 2012 & 57.9\% & 50\\ \hline
  \textbf{YOLOv4} (2020) & CSPDarknet-53 is the backbone of YOLOv4, it is the improved version of Darknet-53. Cross-Stage Partial (CSP) connections are used for better gradient flow and reduced computation. & COCO, PASCAL VOC 2007 and VOC 2012 & COCO, ImageNet & 65.7\% & 62\\ \hline
  \textbf{YOLOv5} (2020) & YOLOv5 is completely written in PyTorch, CSP module is used for feature extraction, FPN and PANet is there for better multi-scale detection & COCO & COCO & 36.7(s),50.7(x) & 156(s), 72(x)\\ \hline
  \textbf{YOLOv6} (2022) & It uses a RepVGG or CSPStackRep-style backbone and BiFPN (Bidirectional Feature Pyramid Network) & COCO & COCO & 35.9(N),52.5(N) & 1231(N), 420(N)\\ \hline
  \textbf{YOLOv7} (2022) & It uses Extended Efficient Layer Aggregation Network (E-ELAN) and also introduced RepConv (Reparameterized Convolution) to improve training. & COCO & COCO & 37.4(Tiny),53.9(X) & 943(Tiny), 114(X)\\ \hline
   \textbf{YOLOv8} (2023) & It uses a RepVGG or CSPStackRep-style backbone and BiFPN (Bidirectional Feature Pyramid Network) & COCO & COCO & 67(n), 87(x) & 700(n), 100(x)\\ \hline
  \end{tabularx}
  \caption{A Comparative Analysis of the YOLO Models}
\end{table}

\subsection{SSD}
The Single Shot MultiBox Detector (SSD), introduced in 2016 by Wei Liu et. al.[58], is an innovative deep learning model designed for real-time object detection. Unlike traditional two-stage approaches like Faster R-CNN, SSD adopts a single-stage architecture that integrates feature extraction, object classification, and localization into a unified framework. By eliminating the separate region proposal step, SSD significantly boosts detection speed while maintaining strong accuracy. Leveraging multi-scale feature maps and predefined default boxes, it efficiently detects objects of various sizes and aspect ratios in a single forward pass. With its lightweight design and real-time processing capabilities, SSD is well-suited for applications such as autonomous vehicles, surveillance, robotics, and mobile devices. Its efficiency and adaptability have made it a standard in object detection, enabling faster and more scalable solutions across multiple domains.

\begin{itemize}
  \item \textbf{Architecture}
  The Single Shot MultiBox Detector (SSD) is designed for fast and accurate object detection by seamlessly integrating feature extraction, classification, and localization into a single framework. At its core, SSD utilizes a pre-trained Convolutional Neural Network (CNN) backbone, such as VGG16 or MobileNet, to analyze input images and generate feature maps rich in spatial and semantic information. A key innovation of SSD is its use of multi-scale feature maps, which enable the detection of objects of various sizes by leveraging different network layers. Higher-resolution layers capture small objects, while deeper layers with a broader receptive field detect larger ones.\\
  To facilitate precise localization, SSD employs default (anchor) boxes at each position on the feature maps, with predefined sizes and aspect ratios. For each default box, SSD predicts class probabilities and bounding box adjustments to refine object localization. Post-processing techniques, such as Non-Maximum Suppression (NMS), filter overlapping predictions, ensuring only the most confident detections remain.\\
  Thanks to its efficient architecture, SSD achieves real-time object detection, making it highly suitable for applications in autonomous vehicles, surveillance, robotics, and mobile devices, where speed and accuracy are critical.
\end{itemize}

  \begin{figure}[ht]
  \centering
  \includegraphics[width=0.9\textwidth, height=3in]{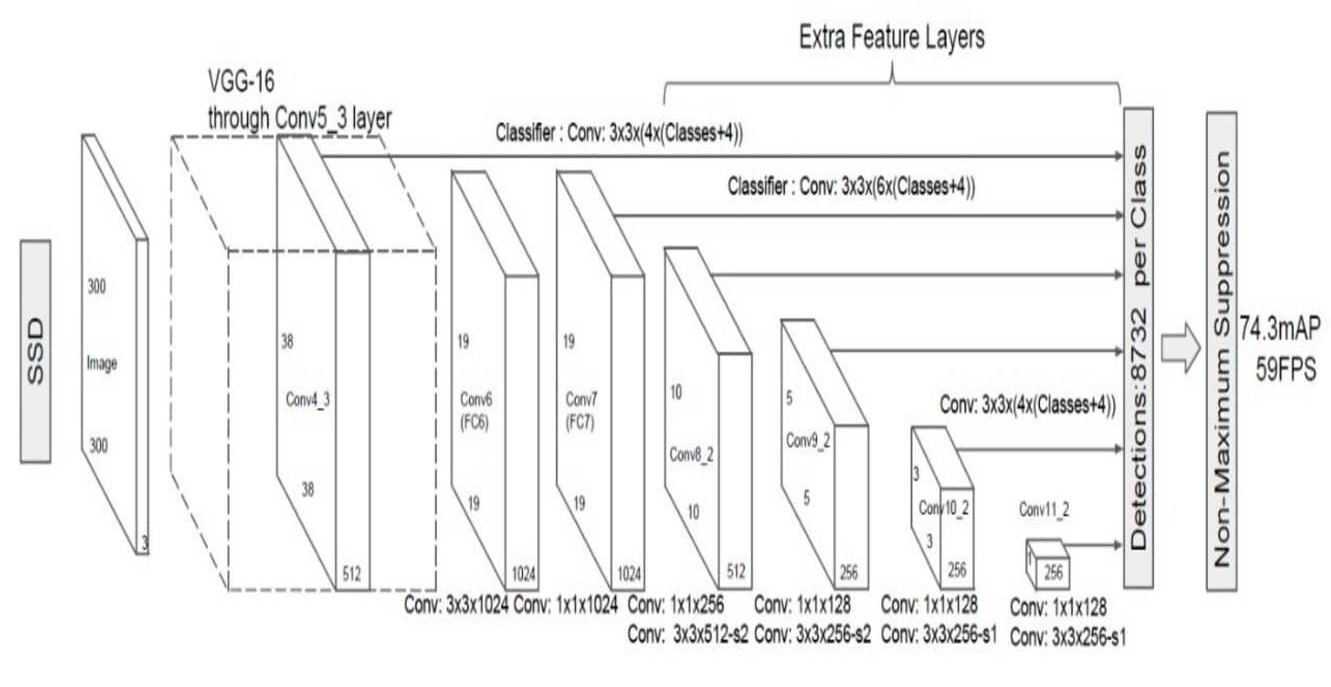}
  \caption{ Architecture of SSD}
  \end{figure}

\subsection{RETINA NET}
Introduced in 2017 by Facebook AI Research, RetinaNet [17] is a powerful object detection model designed to tackle the challenge of class imbalance in images. Traditional detectors often struggle when the background dominates the scene, making it difficult to detect smaller or less frequent objects. RetinaNet addresses this issue with Focal Loss, a novel loss function that reduces the influence of easily classified examples and prioritizes harder-to-detect instances. This approach enhances detection accuracy without compromising efficiency.
Unlike two-stage detectors like Faster R-CNN, RetinaNet operates as a single-stage detector, achieving a balance between speed and performance. It leverages a Feature Pyramid Network (FPN)[59] to improve multi-scale detection, enabling it to recognize objects of varying sizes effectively. Due to its accuracy and efficiency, RetinaNet is widely used in real-time applications, including autonomous vehicles, security systems, robotics, and medical imaging.

\begin{itemize}
  \item \textbf{Architecture}
  RetinaNet's architecture is designed to tackle object detection challenges, particularly class imbalance, by integrating Focal Loss. It consists of three key components: a backbone network for feature extraction, a Feature Pyramid Network (FPN) for multi-scale detection, and two subnetworks for classification and bounding box regression.The backbone, typically a pre-trained CNN like ResNet or ResNeXt, processes the input image to generate feature maps rich in spatial and semantic information. These feature maps are then passed through the FPN, which fuses outputs from different layers of the network, enabling the detection of objects across various scales. This multi-scale approach improves RetinaNet's ability to identify both small and large objects effectively. At each location in the feature maps, RetinaNet generates anchor boxes with predefined sizes and aspect ratios to accommodate objects of varying shapes. It employs two specialized sub-networks: one for classifying anchor boxes and estimating object category probabilities, and another for refining box positions by predicting bounding box adjustments. A key innovation in RetinaNet is the use of Focal Loss, which mitigates class imbalance by directing the model’s focus toward challenging detections while reducing the influence of easily classified background elements. To refine its predictions, Non-Maximum Suppression (NMS) is applied to eliminate redundant boxes, ensuring that only the most confident detections are retained. This efficient design enables RetinaNet to achieve both speed and accuracy, making it well-suited for real-time applications such as autonomous vehicles, surveillance, and robotics.
\end{itemize}  

  \begin{figure}[ht]
  \centering
  \includegraphics[width=0.9\textwidth, height=3in]{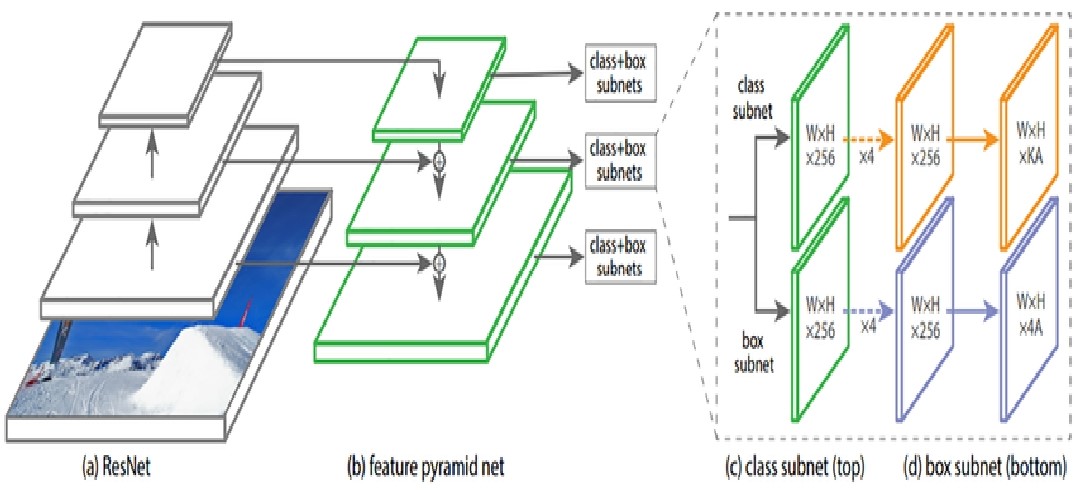}
  \caption{ Architecture of RetinaNet}
  \end{figure}

\subsection{CenterNet}
CenterNet, introduced by Zhou et al. in 2019 [18], is a revolutionary deep learning framework for object detection. Unlike traditional models that rely on bounding-box proposals or anchor boxes, CenterNet adopts a simpler and more efficient anchor-free approach by representing objects as their center points. By identifying the center of each object and predicting its associated attributes, such as size and dimensions, CenterNet significantly reduces computational complexity while maintaining high accuracy.
This streamlined method has transformed object detection, improving both speed and adaptability, making it particularly effective for real-time applications like autonomous vehicles, robotics, and surveillance systems. Its efficiency and versatility have established CenterNet as a preferred choice for modern detection tasks, including 2D and 3D object detection as well as pose estimation. By prioritizing simplicity without sacrificing performance, CenterNet has paved the way for a new generation of object detection models.

\begin{itemize}
  \item \textbf{Architecture}
  CenterNet streamlines object detection by identifying objects as central points rather than relying on anchor boxes or region proposals. It begins with a backbone network, such as ResNet [47] or Hourglass, which extracts feature maps rich in spatial and semantic information from the input image. These feature maps serve as the foundation for predictions, which are generated by the detection head in three key components: a heatmap to indicate object centers, a size regression to estimate the height and width of bounding boxes, and an offset regression to refine the object's exact center position at the pixel level. The heatmap highlights object centers using Gaussian peaks, where intensity represents detection confidence. The size regression predicts bounding box dimensions, while the offset regression ensures precise localization.To enhance training efficiency, CenterNet employs focal loss for heatmap learning to address class imbalance and L1 loss for predicting size and offset values. After predictions are made, post-processing techniques like Non-Maximum Suppression (NMS) help remove duplicate detections and finalize the bounding boxes. With its single-stage, anchor-free design, CenterNet is highly efficient and scalable, making it well-suited for real-time applications such as 2D and 3D object detection and pose estimation, all while maintaining strong accuracy.
\end{itemize}

  \begin{figure}[t]
  \centering
  \includegraphics[width=0.9\textwidth, height=4in]{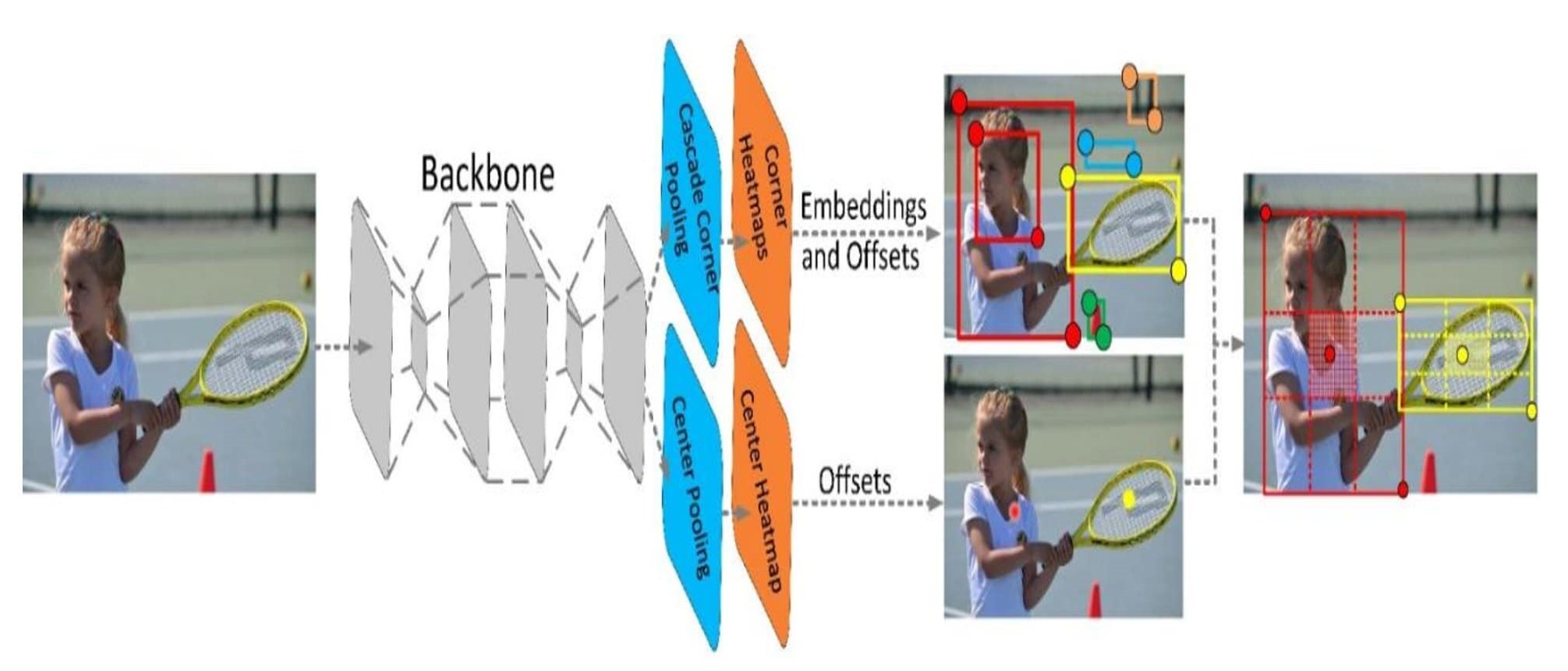}
  \caption{ Architecture of CenterNet}
  \end{figure}

\subsection{Efficient Det}
EfficientDet is a state-of-the-art object detection model introduced in 2020 by Mingxing Tan, Ruoming Pang, and Quoc V [60]. Le in their paper, "EfficientDet: Scalable and Efficient Object Detection." Designed to balance accuracy and computational efficiency, it is particularly well-suited for resource-constrained environments. Built on the EfficientNet backbone for feature extraction, EfficientDet introduces two key innovations: the Bi-directional Feature Pyramid Network (BiFPN) for optimized multi-scale feature fusion and a novel scaling strategy that simultaneously adjusts resolution, depth, and width. These enhancements allow EfficientDet to outperform earlier models like RetinaNet and Faster R-CNN in both speed and accuracy, all while significantly reducing parameter count and computational cost (FLOPs).
The EfficientDet model family is highly adaptable, ranging from lightweight versions optimized for mobile applications [61][62] to high-performance variants designed for server-level tasks [63]. Its ability to deliver accurate detections with minimal computational demand has made it a popular choice in fields such as autonomous driving, surveillance, and medical imaging. By prioritizing efficiency and scalability, EfficientDet has set a new standard for modern object detection systems, demonstrating that high precision can be achieved without excessive resource consumption.

\begin{itemize}
  \item \textbf{Architecture}
  EfficientDet's architecture is designed to balance computational efficiency and high accuracy in object detection. Built on the EfficientNet model series, it benefits from a streamlined and optimized design. EfficientNet employs a compound scaling method, which adjusts resolution, depth, and width simultaneously to maximize performance while keeping computational costs low.\\
  A key innovation in EfficientDet is the Bi-directional Feature Pyramid Network (BiFPN), which enhances multi-scale feature integration. Unlike traditional feature pyramids, BiFPN enables two-way information flow across feature levels, allowing features from different scales to be combined more effectively. Additionally, it incorporates trainable weights, prioritizing the most important features during training, which improves efficiency while reducing computational overhead.\\
  EfficientDet's detection head processes the combined features to generate object classification and bounding box regression predictions, maintaining a lightweight design for improved efficiency. The model leverages compound scaling, enabling it to scale from EfficientDet-D0 (a compact variant) to EfficientDet-D7 (a high-performance version), ensuring adaptability across different hardware and performance requirements.\\
  During training, EfficientDet employs Focal Loss to address class imbalance and Smooth L1 Loss for precise bounding box regression. These features make EfficientDet highly scalable, efficient, and accurate, delivering state-of-the-art performance while minimizing computational demands. Its adaptability makes it an ideal choice for both mobile applications and high-performance computing environments.

  \item \textbf{Variations}
  EfficientDet comes in multiple versions, each designed to accommodate different resource constraints and performance demands. The models, ranging from EfficientDet-D0 to EfficientDet-D7, represent varying levels of complexity and computational efficiency. EfficientDet-D0 is the smallest and fastest variant, optimized for mobile devices and low-power environments, making it ideal for real-time object detection in applications such as mobile cameras and embedded systems.\\
  As the model number increases, so does its complexity and capability. EfficientDet-D1 to D3 strike a balance between accuracy and computational cost, making them well-suited for edge devices and moderate-performance applications such as drones and smart cameras [64][65]. The larger variants, EfficientDet-D4 to D7, are designed for high-performance tasks where precision is paramount, including autonomous vehicles, surveillance systems, and large-scale image analysis.\\
  All versions of EfficientDet share the same core architecture, utilizing compound scaling to adjust resolution, depth, and width based on the model variant. This adaptive scaling ensures that each version operates at optimal efficiency for its intended application, allowing EfficientDet to be deployed seamlessly across a wide range of environments—from resource-constrained mobile devices to high-performance cloud-based platforms.
\end{itemize}

  \begin{figure}[t]
  \centering
  \includegraphics[width=0.9\textwidth, height=4in]{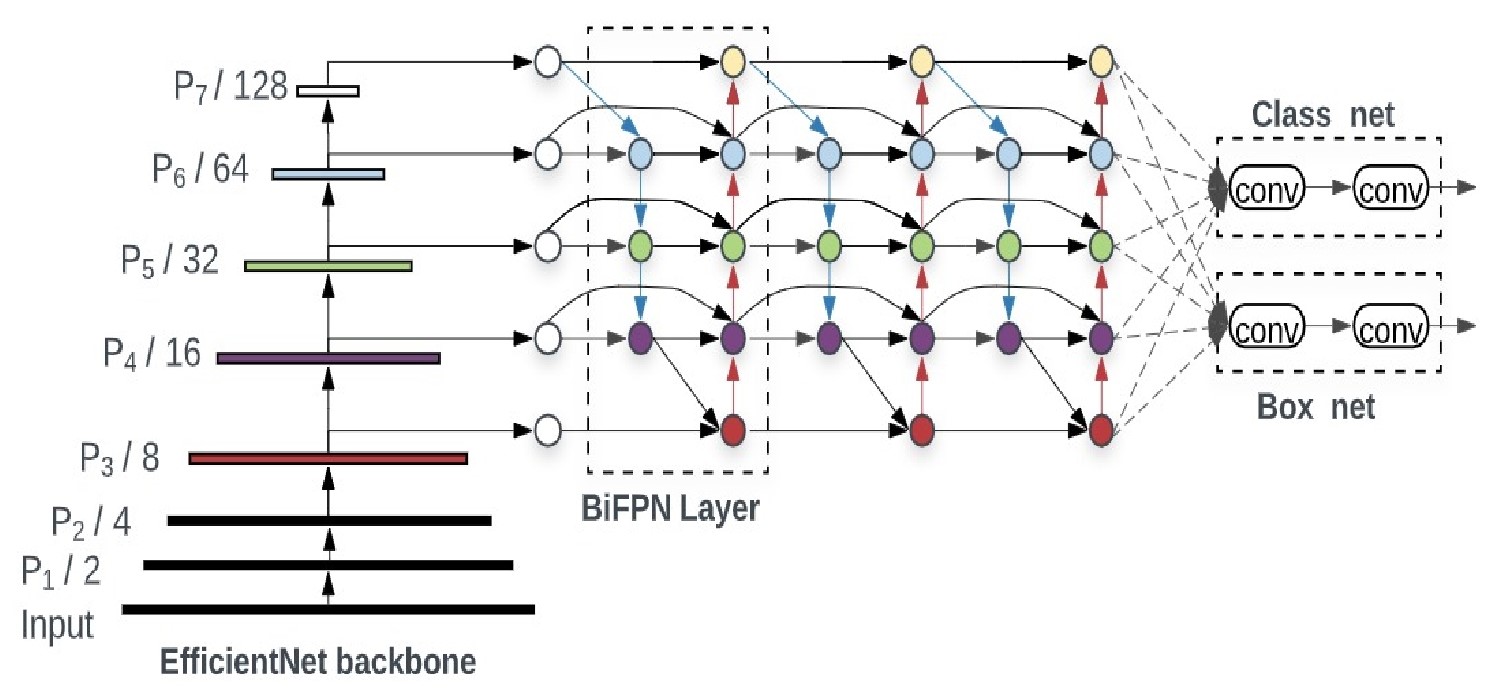}
  \caption{ Architecture of EfficientDet}
  \end{figure}

\subsection{Lighter Head R-CNN}
The Lighter Head R-CNN is an improved version of the traditional R-CNN model, designed to enhance efficiency and reduce computational costs while maintaining high accuracy in object detection. It optimizes the head architecture responsible for processing region proposals for classification and bounding box adjustments by incorporating efficient design elements like separable convolutions or compact layers. Inspired by the work of Zeming Li et al. in 2017 [66], this model aims to refine two-stage object detectors like Faster R-CNN while retaining their superior accuracy compared to single-stage detectors such as YOLO and SSD.
Compared to earlier models, Lighter Head R-CNN offers significant advantages. While CNNs are effective for feature extraction and image classification, they are not inherently designed for direct object detection. Traditional R-CNN, though highly accurate, is slow due to its region-based approach. Fast R-CNN improved efficiency with ROI pooling but remained limited by its dependence on selective search for region proposals. Faster R-CNN addressed this by introducing Region Proposal Networks (RPNs), which enabled integrated training and enhanced both speed and accuracy, though its detection head remained computationally expensive. In contrast, single-stage detectors like YOLO and SSD focus on speed but often compromise accuracy.
Lighter Head R-CNN bridges this gap by incorporating an efficient detection head, achieving faster inference times than Faster R-CNN while maintaining higher accuracy compared to YOLO and SSD. It retains the strengths of two-stage detectors, excelling in detecting small objects and handling complex scenes, while also being well-suited for large datasets and high-resolution images[67]. Its adaptability allows deployment across diverse hardware platforms, from powerful GPUs to energy-efficient CPUs. This makes Lighter Head R-CNN an ideal choice for applications that demand a balance of speed, efficiency, and reliable performance in real-world scenarios.

\begin{itemize}
  \item \textbf{Architecture}
  Lighter Head R-CNN fills this gap by integrating a highly efficient detection head, delivering faster inference than Faster R-CNN while outperforming YOLO and SSD in accuracy. It leverages the advantages of two-stage detectors, excelling at detecting small objects and managing complex scenes, making it well-suited for large datasets and high-resolution images. Its versatility enables seamless deployment across various hardware platforms, from high-performance GPUs to energy-efficient CPUs. This makes Lighter Head R-CNN a top choice for applications requiring a balance of speed, efficiency, and dependable performance in real-world environments.
  The model employs ROI Align to precisely align region proposals with the feature map, ensuring accurate extraction of fixed-size features for each region. A lightweight detection head then processes these features to classify objects and refine bounding boxes. Object classification is performed using Cross-Entropy Loss, while Smooth L1 Loss is used for bounding box regression. By sharing features across tasks and optimizing the head structure, the model significantly reduces computational overhead, making it more efficient than traditional Faster R-CNN. Its streamlined architecture enables it to handle high-resolution images and large datasets efficiently, making it well-suited for real-time applications and resource-constrained environments. Overall, Lighter Head R-CNN strikes a balance between processing efficiency and detection accuracy, making it a versatile choice for various object detection tasks[68][69][70].
\end{itemize}

  \begin{figure}[t]
  \centering
  \includegraphics[width=0.9\textwidth, height=4in]{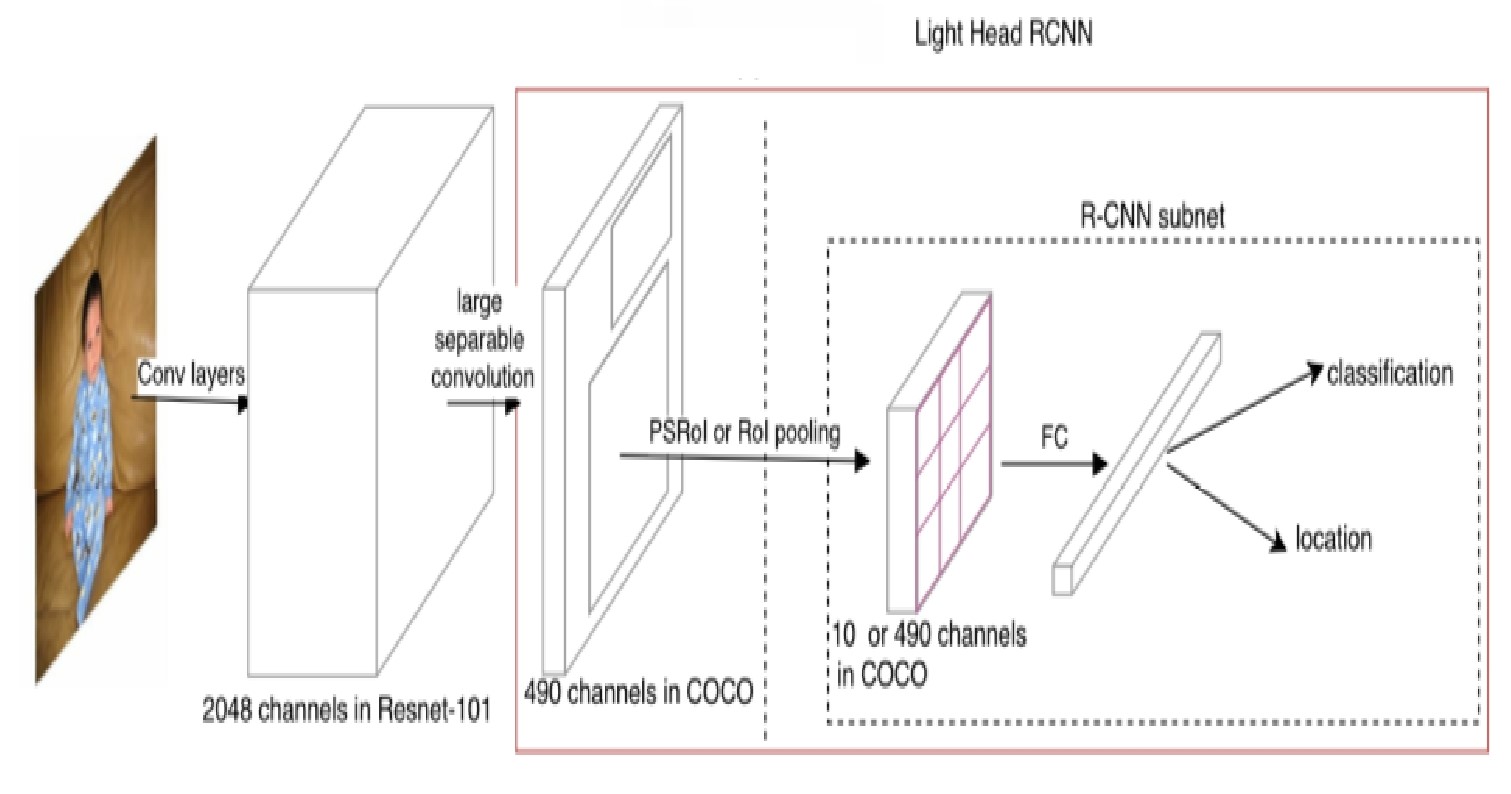}
  \caption{ Architecture of Light Head RCNN}
  \end{figure}

  \begin{figure}[t]
  \centering
  \includegraphics[width=1\textwidth, height=7in]{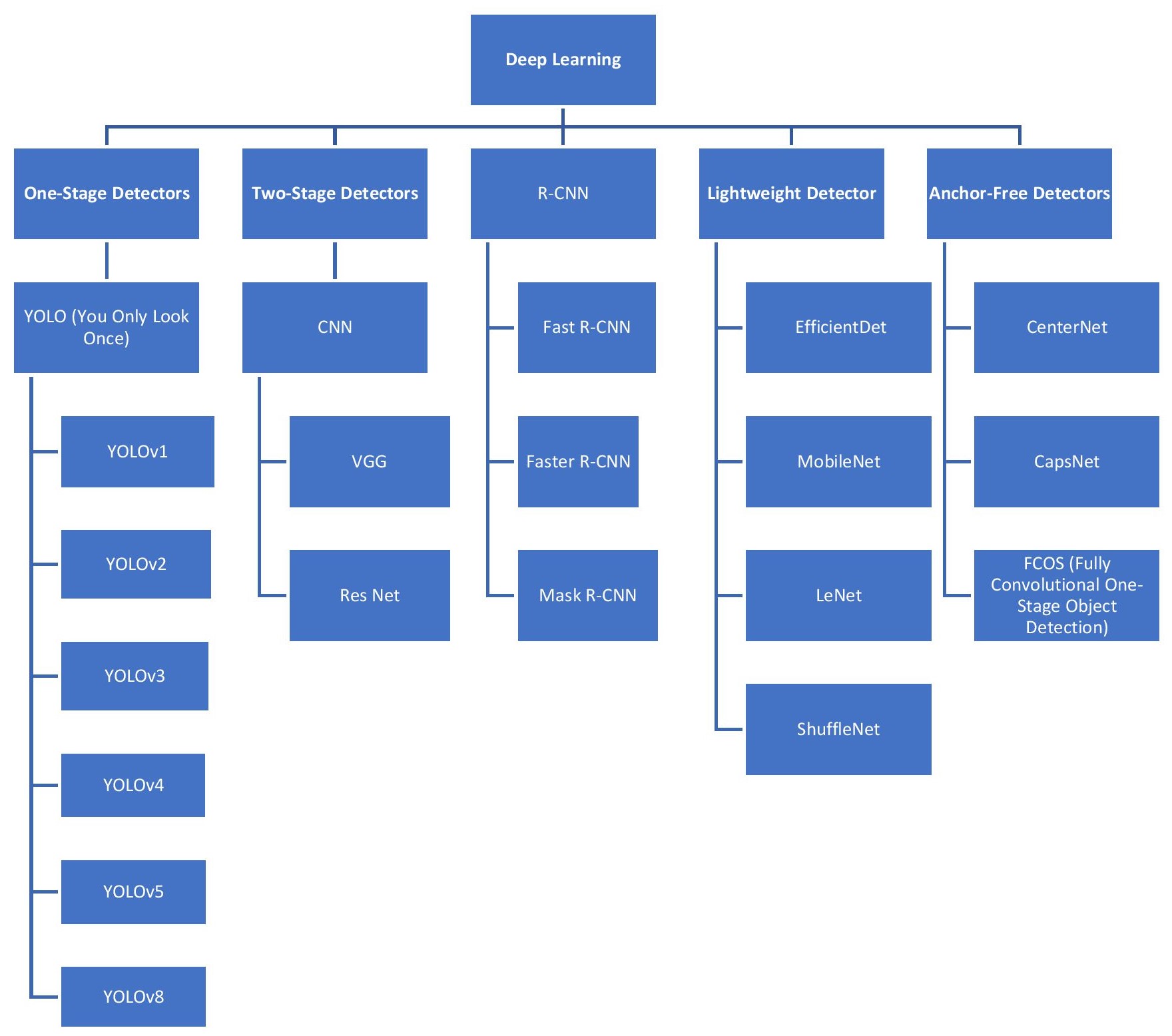}
  \caption{Classification of deep learning algorithms into one-stage detectors, two-stage detectors, lightweight detectors, and anchor-free detectors.}
  \end{figure}  

\clearpage

\begin{table}[ht]
  \scriptsize
  \centering
  \begin{tabularx}{\textwidth}{|X|X|X|X|X|}
  \hline
  \textbf{Dataset} & \textbf{Description} & \textbf{Type of Images} & \textbf{Dataset Source} \\ \hline
   DUTS & DUTS is a large saliency detection dataset with 10,553 training and 5,019 testing images, sourced from ImageNet DET and SUN datasets. It includes challenging cases, with pixel-level annotations by 50 individuals. & Living and non-living species & \href{https://www.kaggle.com/datasets/balraj98/duts-saliency-detection-dataset}{Kaggle}\\ \hline
   Car Detection & The dataset contains media of cars in all views. Here we see 1001 images for training and 175 for testing & cars & \href{https://www.kaggle.com/datasets/sshikamaru/car-object-detection}{Kaggle}\\ \hline
   Gun Detection & Images folder contains pictures in jpeg format. Labels folder holds txt files where each file's first line shows the number of items in the matching image, and the following lines include the coordinates of the box outlining the item. & Guns & \href{https://www.kaggle.com/datasets/issaisasank/guns-object-detection}{Kaggle}\\ \hline
   Human Face detection & A varied collection of images showing human faces from different races, ages, and angles, intended to build an impartial dataset that contains coordinates of facial areas suitable for training models that detect objects. & Human Faces  & \href{https://www.kaggle.com/datasets/sbaghbidi/human-faces-object-detection}{Kaggle}\\ \hline
   Brain tumor detection  & The datasets JPGs exported at their native size and are separated by plane (Axial, Coronal and Sagittal). Here we see 310 images for training and 75 for testing & Brain Tumor & \href{https://www.kaggle.com/datasets/davidbroberts/brain-tumor-object-detection-datasets}{Kaggle}\\ \hline
   26 Class Object Detection Dataset & This Dataset contains labeled images of 26 common city and outdoor objects, including vehicles, traffic signs, pedestrians, and natural elements. & City elements & \href{https://www.kaggle.com/datasets/mohamedgobara/26-class-object-detection-dataset}{Kaggle}\\ \hline
   Underwater Object Detection & The dataset includes seven underwater creature classes with bounding box annotations. It is pre-split into training (448 images), validation (147 images), and test (63 images) sets. & Under water species & \href{https://www.kaggle.com/datasets/slavkoprytula/aquarium-data-cots}{Kaggle}\\ \hline
  Raccoon Detection Dataset & There are 213 bounding boxes and 196 raccoon photos in this collection. The pictures vary in size, and this is a one class problem. It is a great starting dataset to start with object detection. & Raccoon & \href{https://public.roboflow.com/object-detection/raccoon/2}{Roboflow}\\ \hline
  Poribohon-BD & The Poribohon-BD dataset contains 9,058 images of 15 Bangladeshi vehicle types, captured under diverse poses, angles, lighting, and weather conditions. & vehicles & \href{https://data.mendeley.com/datasets/pwyyg8zmk5/2}{Mendeley}\\ \hline
  Plastic Object Detection  & Images of variety of plastic items that are frequently encountered in daily life are included in this dataset. Bounding boxes encircle the plastic objects in each photograph & Plastic Objects  & \href{https://www.kaggle.com/datasets/dataclusterlabs/plastic-object-detection-dataset}{Kaggle}\\ \hline
  Traffic vehicles Detection & Traffic vehicle detection identifies and locates vehicles in images or videos for applications like surveillance, traffic control, autonomous driving, and intelligent transportation systems. & Various vehicle types in road images or video streams.  & \href{https://gts.ai/dataset-download/traffic-vehicles-premier-object-detection-dataset}{GTS}\\ \hline
  WordNet-ImageNet hierarchy & It refers to the structured organization of object categories in ImageNet, which is built upon WordNet. ImageNet dataset has 1 lakh 40 thousand images and 21 thousand four eighty-one no of classes & Images of Plant, Animal and Activity  & \href{https://archive.ics.uci.edu/dataset/693/imagenet}{UCI}\\ \hline
  COCO(Common Objects in Context) & The COCO (Common Objects in Context) dataset is widely used for object detection, segmentation (instance and panoptic), keypoint detection, and captioning tasks. It contains over 330,000 images, with 1.5 million object instances spanning 80 object categories, making it a benchmark dataset for deep learning-based computer vision tasks. & COCO contains diverse real-world images with objects, scenes, crowds, actions, varied lighting, and annotations.  & \href{https://cocodataset.org}{COCO-HOME}\\ \hline
  PASCAL VOC & The PASCAL Visual Object Classes (PASCAL VOC) dataset consists of standardized visual datasets collected through VOC challenges, designed for object detection, segmentation, and classification in real-world images. & It has 11,530 no. of images and 20 classes  & \href{http://host.robots.ox.ac.uk/pascal/VOC/}{GTS}\\ \hline
  \end{tabularx}
\end{table}

\clearpage

\section{Different application domains for real-time object detection.}
\subsection{Generic object detection:}
A key challenge in computer vision is achieving accurate generic object detection, which involves locating and categorizing objects in images or videos. This is essential for applications like image segmentation, autonomous driving, robotics, and video surveillance. Object detection is broadly classified into region-based methods (R-CNN family)[71][72][73] and single-shot detectors [74] (e.g.YOLO). Multi-stage detection systems often face challenges due to their sequential prediction of object locations and categories [75].

Traditional object detection methods relied on RANSAC techniques for image matching using hand-crafted features like Histogram of Oriented Gradients (HOG)[76] and Scale-Invariant Feature Transform (SIFT)[77]. These approaches combined feature recognition with Support Vector Machines (SVMs)[78][79] but struggled with variations in lighting, background complexity, and spatial constraints. The introduction of Convolutional Neural Networks (CNNs) marked a major advancement, enabling both feature extraction and multi-scale representation. Modern two-stage detectors, such as R-CNN, revolutionized object detection by leveraging CNNs. Faster R-CNN further improved efficiency by integrating Region Proposal Networks (RPN) for seamless region generation. Subsequent enhancements, like Mask R-CNN [80], added pixel-level instance segmentation, while Cascade R-CNN [81] introduced a multi-stage refinement process, improving performance at higher Intersection over Union (IoU) thresholds. These advancements have significantly outperformed traditional methodologies, establishing deep learning-based approaches as the standard for object detection.

Single-stage detectors like YOLO and SSD improve efficiency by predicting class probabilities and bounding boxes in one step. YOLO versions have evolved for better speed and accuracy, while RetinaNet uses focal loss to handle class imbalance. EfficientDet optimizes feature pyramids, and transformer-based models like DETR [82] and Deformable DETR enhance detection efficiency. Multiscale feature integration enhances object detection, with models like FPN [83] building feature hierarchies and HRNet[84] preserving high resolution for better accuracy. TridentNet [85] improves multiscale learning, while semi-supervised methods like pseudo-labeling and STAC reduce reliance on labeled data [86]. Performance is assessed using datasets like COCO and Pascal VOC, with mAP and IoU serving as key evaluation metrics. Challenges remain in detecting small or occluded objects, real-time processing, and adversarial robustness. Future improvements may include self-supervised learning, multimodal inputs, graph-based modeling, and neural architecture search for increased accuracy and efficiency.

\begin{table}[ht]
  \scriptsize
  \centering
  \begin{tabularx}{\textwidth}{|X|X|X|X|X|}
  \hline
  \textbf{Algorithm} & \textbf{mAP(\%)} & \textbf{IoU} &\textbf{Inference Time(ms)} &\textbf{Model Size (MB)} \\ \hline
  \textbf{YOLOv4} & 91.3 & 88.5 & 22 & 250 \\ \hline
  \textbf{Faster R-CNN} & 93.5 & 90.8 & 120 & 500 \\ \hline
  \textbf{SSD} & 87.4 & 83.2 & 18 & 95 \\ \hline
  \textbf{RetinaNet} & 90.8 & 88.0 & 30 & 120 \\ \hline
  \textbf{DETR} & 89.6 & 87.5 & 45 & 150 \\ \hline
  \textbf{EfficientDet} & 92.1 & 89.2 & 25 & 75 \\ \hline
  \textbf{Cascade R-CNN} & 94.2 & 91.5 & 130 & 600 \\ \hline
  \end{tabularx}
  \caption{A comparative analysis of evaluation metrices for COCO dataset}
  \end{table}

\begin{figure}[ht]
  \centering
  \includegraphics[width=0.9\textwidth, height=2in]{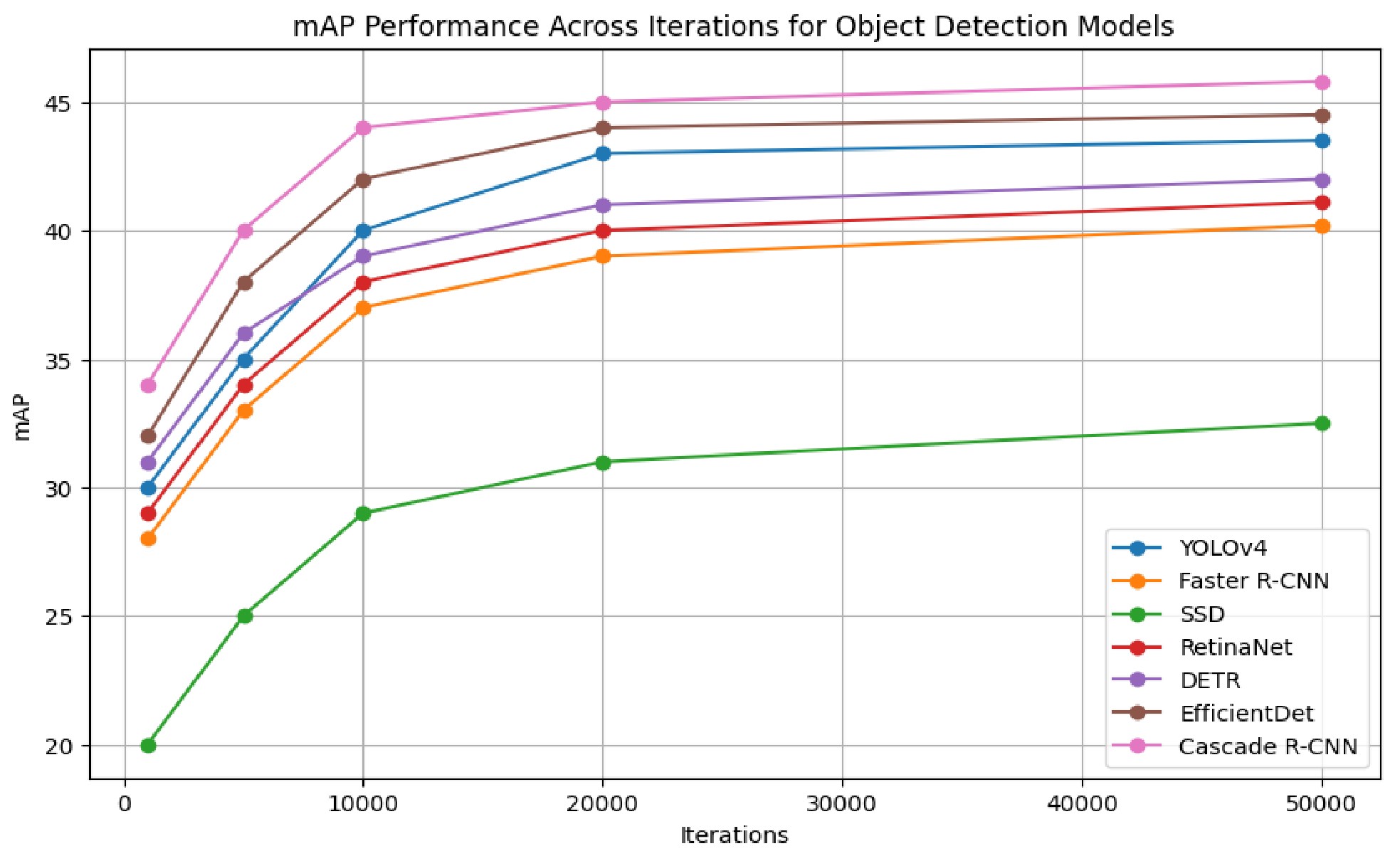}
  \caption{ A comparative analysis of evaluation metrices of object detection models
for COCO dataset}
\end{figure}

\clearpage

\subsubsection{Image classification}
Image classification is a key computer vision task that assigns labels to images. Deep learning has significantly improved classification accuracy, starting with AlexNet [10], which won the 2012 ImageNet competition. VGGNet [87] introduced deeper architectures with smaller filters, while GoogLeNet [46] optimized computation using the inception module. ResNet [47] revolutionized deep learning with residual connections, allowing for extremely deep networks. DenseNet [52] further enhanced efficiency by enabling each layer to receive inputs from all previous layers, promoting feature reuse while reducing parameters. These advancements continue to refine image classification models for improved accuracy and efficiency.

Recent advancements in image classification have integrated attention mechanisms [88][89], leading to models like Vision Transformers (ViT) [90] that replace convolutions with self-attention for global context modeling. Hybrid architectures, such as Swin Transformer [91] and ConViT [92], combine CNNs with transformers, while EfficientNet [56] optimizes performance using compound scaling with minimal parameters. Pretrained models on large datasets like ImageNet facilitate transfer learning, while techniques like data augmentation, dropout, and regularization enhance model generalization. To evaluate image classifiers, key metrics include accuracy, precision, recall, and F1-score, which assess classification effectiveness. Precision measures correct positive predictions, recall quantifies true positive retrieval, and F1-score balances both. Macro- and micro-averaged metrics handle class imbalances, while a confusion matrix visualizes classification errors. For imbalanced or large-scale tasks, AUC-ROC, Average Precision (AP), and top-k accuracy offer additional insights. Calibration metrics like Expected Calibration Error (ECE) assess the reliability of confidence scores. These advancements have made deep learning-based classification standard across applications such as medical imaging [93][94], self-driving cars [95][96], and facial recognition [97][98], expanding its impact across industries [99][100].

\begin{table}[ht]
  \scriptsize
  \centering
  \begin{tabularx}{\textwidth}{|X|X|X|X|X|X|X|}
  \hline
  \textbf{Algorithm} & \textbf{Accuracy(\%)} & \textbf{Precision(\%)} &\textbf{Recall(\%)} &\textbf{F1-Score(\%)} &\textbf{Inference Time(ms)} &\textbf{Model Size(MB)} \\ \hline
  \textbf{Custom CNN} & 80.5 & 79.8 & 80.1 & 80.0 & 5 & 10 \\ \hline
  \textbf{Alex Net} & 85.3 & 84.5 & 84.8 & 84.6 & 20 & 230 \\ \hline
  \textbf{VGG} & 92.4 & 92.0 & 92.2 & 92.1 & 35 & 528 \\ \hline
  \textbf{RestNet} & 94.5 & 94.2 & 94.4 & 94.3 & 25 & 98 \\ \hline
  \textbf{MobileNet} & 89.7 & 89.3 & 89.5 & 89.4 & 10 & 14 \\ \hline
  \textbf{EfficientDet} & 95.1 & 94.8 & 94.9 & 94.9 & 20 & 36 \\ \hline
  \end{tabularx}
  \caption{A comparative analysis of evaluation metrices for CIFAR-10 dataset}
\end{table}

\begin{figure}[ht]
  \centering
  \includegraphics[width=0.9\textwidth, height=3in]{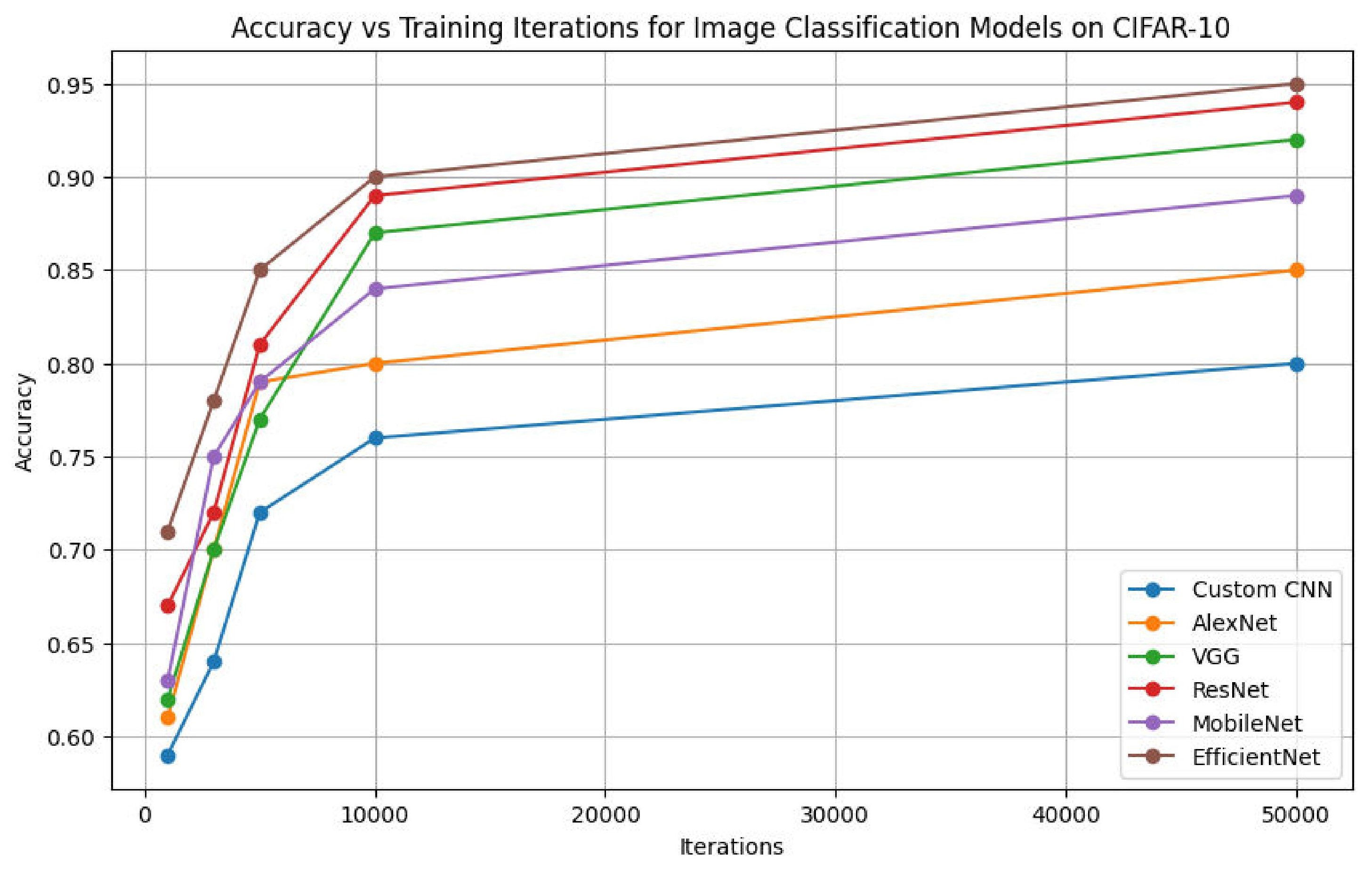}
  \caption{A comparative analysis of evaluation metrices of image classification models for CIFAR-10 dataset}
\end{figure}

\clearpage

\subsubsection{Pedestrian Detection}
Human figure detection plays a crucial role in computer vision, supporting applications like surveillance [101][102], robotics [103][104], and autonomous vehicles [105][106]. Deep learning advancements have significantly improved detection accuracy, efficiency, and real-time processing. Region-based methods such as R-CNN introduced selective search for generating region proposals, which Fast R-CNN and Faster R-CNN later optimized by reducing redundancy and improving processing speed. Region Proposal Networks (RPN) and M-Blox further refined these approaches by lowering computational costs while enhancing accuracy. For real-time detection, SSD provided a fast and efficient architecture, while YOLO revolutionized object detection by framing it as a regression problem, dramatically improving both speed and accuracy. These innovations have enabled real-time human detection, making it more practical for real-world applications. Further advancements, including transformer-based vision models [107][108] and hybrid deep learning architectures, continue to enhance human detection accuracy, robustness, and efficiency, contributing to the broader evolution of computer vision technologies in diverse industries.

Deformable Convolutional Networks enhance pedestrian detection by effectively handling various poses [109]. DeepFusion integrates multiple architectures for improved robustness in challenging scenarios [110][111], while Feature Pyramid Networks enable multi-scale detection, aiding in distance and size recognition.Cascade R-CNN, with its multi-stage refinement, enhances detection in crowded environments, while RetinaNet, utilizing focal loss, performs well in biased and non-biased situations, excelling at detecting small or occluded pedestrians.Transformers have revolutionized pedestrian detection, with the Adaptive Attention Mechanism further refines DETR by focusing on significant inputs, enhancing detection accuracy [112]. Additionally, integrating transformer modules into convolutional backbones has boosted the speed and accuracy of YOLOv4 and YOLOv5 in real-time pedestrian detection [113]. These advancements collectively improve pedestrian detection efficiency, making it more reliable across different conditions and environments.

\begin{table}[ht]
  \scriptsize
  \centering
  \begin{tabularx}{\textwidth}{|X|X|X|X|X|}
  \hline
  \textbf{Algorithm} & \textbf{mAP(\%)} & \textbf{Recall(\%)} &\textbf{Precision(\%)} &\textbf{Inference Time(ms)} \\ \hline
  \textbf{Faster RCNN} & 88.4 & 91.2 & 87.6 & 150 \\ \hline
  \textbf{YOLOv4} & 85.6 & 88.0 & 85.0 & 25 \\ \hline
  \textbf{SSD} & 82.4 & 84.2 & 81.5 & 18 \\ \hline
  \textbf{RetinaNet} & 86.5 & 89.5 & 85.7 & 40 \\ \hline
  \textbf{CenterNet} & 87.3 & 89.8 & 86.4 & 30 \\ \hline
  \textbf{Cascade RCNN} & 90.1 & 92.7 & 89.5 & 200 \\ \hline
  \end{tabularx}
  \caption{A comparative analysis of evaluation metrices for Caltech Pedestrian Dataset}
\end{table}

\begin{figure}[ht]
  \centering
  \includegraphics[width=0.9\textwidth, height=3in]{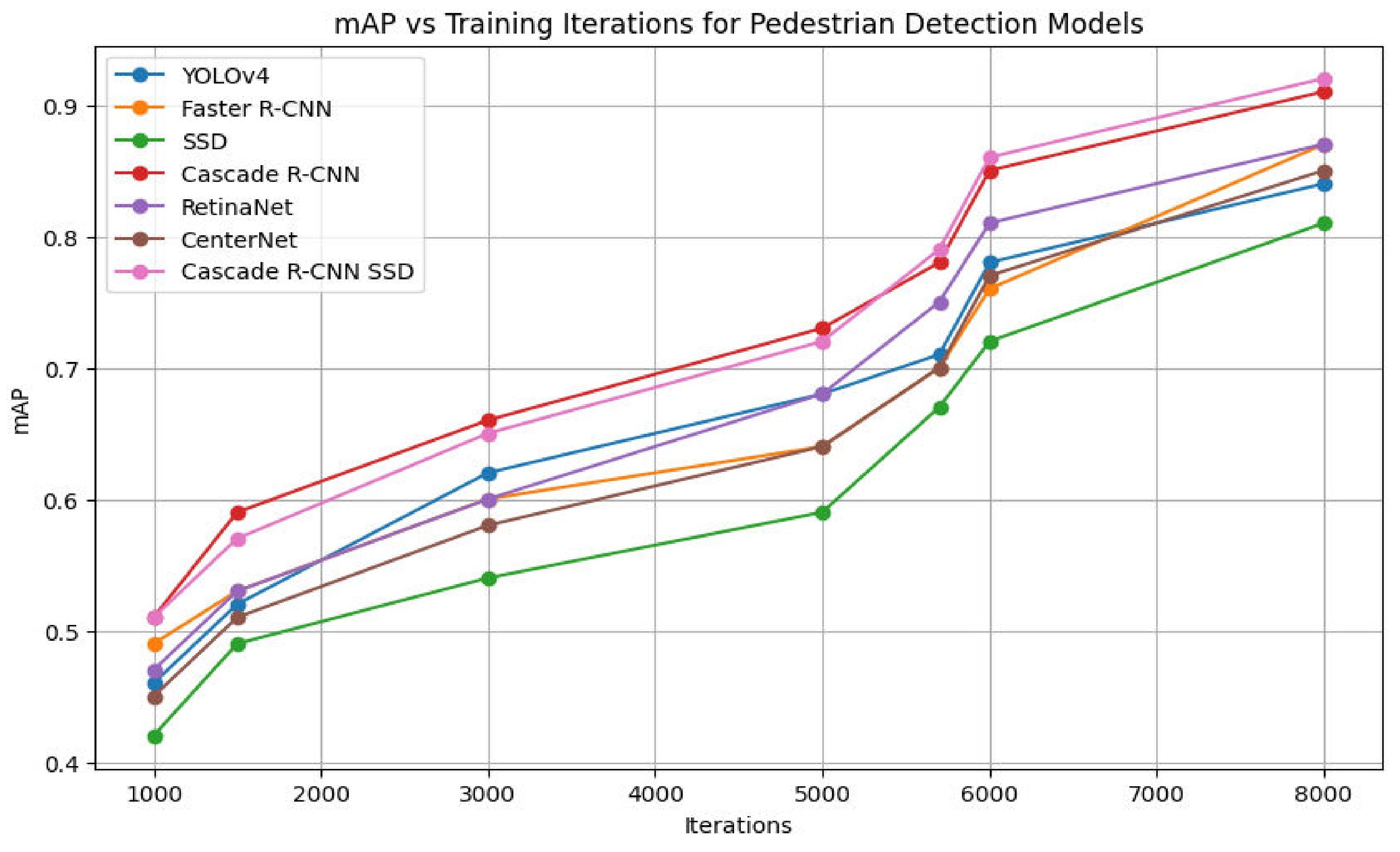}
  \caption{A comparative analysis of evaluation metrices for Caltech Pedestrian Dataset}
\end{figure}

\clearpage

\subsubsection{Skeleton Detection}
Skeleton detection, or human pose estimation [114][115][116], involves identifying body joints like elbows, shoulders, and knees from images or videos. It plays a key role in action recognition [117], healthcare [118], sports [119], animation [120], and human-computer interaction [121]. Deep neural networks have significantly enhanced accuracy by analyzing spatial and temporal dependencies. Early approaches included Convolutional Pose Machines (CPM) [122], which used convolutional layers for joint localization. OpenPose improved this by introducing part affinity fields for multi-person pose estimation. The Stacked Hourglass Network [123] further refined joint detection using a symmetric encoder-decoder architecture. Top-down methods, such as Faster R-CNN, first detect people before identifying keypoints. Mask R-CNN improved this by integrating keypoint detection, while HRNet [124] enhanced accuracy in localizing small or occluded joints by maintaining high-resolution representations. These advancements have led to more precise and real-time pose estimation, making skeleton detection highly effective across various applications.

Neural networks have significantly advanced skeleton detection. PoseFormer leverages self-attention to model joint dependencies, enhancing pose estimation in complex and crowded environments [125]. Integrating pose estimation into end-to-end detection frameworks, such as Dynamic Histogram Symmetry(DHS) and (VIH)[126][127], has shown promising results. Temporal skeleton detection is crucial for video analysis. ST-GCN applies graph theory to capture spatial and temporal dependencies in human skeletons [128], while 2s-AGCN refines this by modeling skeleton sequences adaptively. Combining 2D pose estimations with 3D neural networks, as seen in PoseC3D [129], further improves detection accuracy. Multi-scale information enhances skeleton detection precision. DeepPose [130] treats pose estimation as a regression task, while MSPN introduces a multi-stage architecture with cross-stage feature aggregation, significantly improving joint localization accuracy. These advancements continue to refine skeleton detection for real-world applications.

\begin{table}[ht]
  \scriptsize
  \centering
  \begin{tabularx}{\textwidth}{|X|X|X|X|X|}
  \hline
  \textbf{Algorithm} & \textbf{PCK(\%)} & \textbf{mAP(\%)} &\textbf{Inference Time(ms)} &\textbf{Model Size(MB)} \\ \hline
  \textbf{OpenPose} & 85.3 & 80.1 & 35 & 200 \\ \hline
  \textbf{HRNet} & 90.2 & 87.5 & 50 & 220 \\ \hline
  \textbf{PoseNet} & 78.5 & 70.3 & 12 & 50 \\ \hline
  \textbf{DeepPose} & 70.5 & 65.1 & 25 & 150 \\ \hline
  \end{tabularx}
  \caption{A comparative analysis of evaluation metrices for Skeleton Detection models on COCO Dataset}
\end{table}

\begin{figure}[ht]
  \centering
  \includegraphics[width=0.9\textwidth, height=3in]{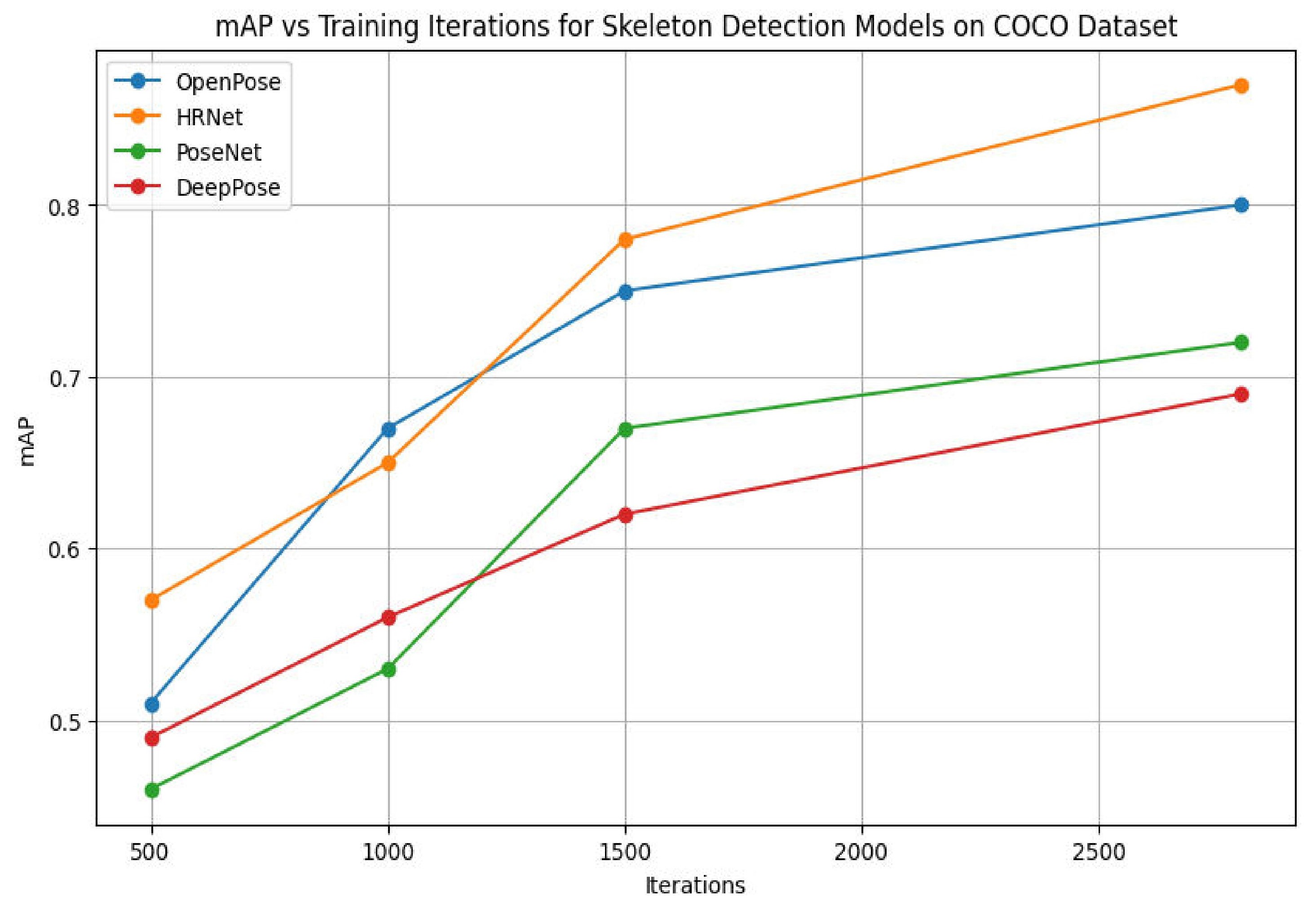}
  \caption{A comparative analysis of evaluation metrices for Skeleton Detection models on COCO Dataset}
\end{figure}

\clearpage

\subsection{Salient Object Detection:}
Salient Object Detection (SOD) is essential in computer vision for identifying and segmenting key objects in images [131], benefiting tasks like image segmentation, object recognition, and visual tracking. Traditional methods relied on handcrafted features such as contour, color, and texture, but deep learning has significantly improved SOD performance [132][133][134]. Convolutional Neural Networks (CNNs) have proven highly effective, enabling end-to-end saliency prediction without handcrafted features. Early deep learning models like Deep Contrast Network (DCL)[135] and Multi-task Deep Saliency Model (MTDS)[136] introduced two-stream architectures and integrated saliency detection with semantic segmentation. Further advancements, such as U-Net [49], enhanced SOD by using skip connections to merge low and high-level features, producing sharper saliency maps [137]. The Pyramid Pooling Network further improved performance by integrating multi-scale contextual information, enabling better detection in complex environments[138]. These innovations have greatly refined SOD, making it more accurate and efficient.

Recent advances in Salient Object Detection (SOD) have been driven by attention mechanisms that focus on key image regions [139][140][141]. Attention U-Net and PiCANet enhance saliency by applying spatial and contextual attention [142], while transformer-based models like ViT-SOD use self-attention for improved accuracy [143]. Multi-scale networks such as MSPFN and ASNet [144] fuse features at different resolutions to detect objects of various sizes [145]. RAS further refines saliency maps using recurrent attention [146]. For video SOD, temporal information is critical. 3D CNNs, ConvLSTM, and Two-Stream Networks capture motion and combine spatial-temporal saliencies for better detection of moving objects.

Evaluation relies on benchmark datasets  like MSRA-B, DUTS, HKU-IS, and ECSSD. DUTS is the largest with over 15,000 images, while MSRA-B includes 5,000 annotated natural images. Common metrics include F-measure, MAE, and S-measure, assessing precision, pixel error, and structural consistency.

\begin{table}[ht]
  \scriptsize
  \centering
  \begin{tabularx}{\textwidth}{|X|X|X|X|X|}
  \hline
  \textbf{Algorithm} & \textbf{F-measure} & \textbf{MAE} &\textbf{S-measure} &\textbf{Inference Time (ms)} \\ \hline
  \textbf{U-Net} & 0.78 & 0.045 & 0.85 & 20 \\ \hline
  \textbf{RFCN} & 0.85 & 0.032 & 0.88 & 25 \\ \hline
  \textbf{PiCANet} & 0.90 & 0.020 & 0.91 & 40 \\ \hline
  \textbf{DSS} & 0.87 & 0.028 & 0.90 & 28 \\ \hline
  \end{tabularx}
  \caption{A comparative analysis of evaluation metrices for DUTS Dataset}
\end{table}

\begin{figure}[ht]
  \centering
  \includegraphics[width=0.9\textwidth, height=3in]{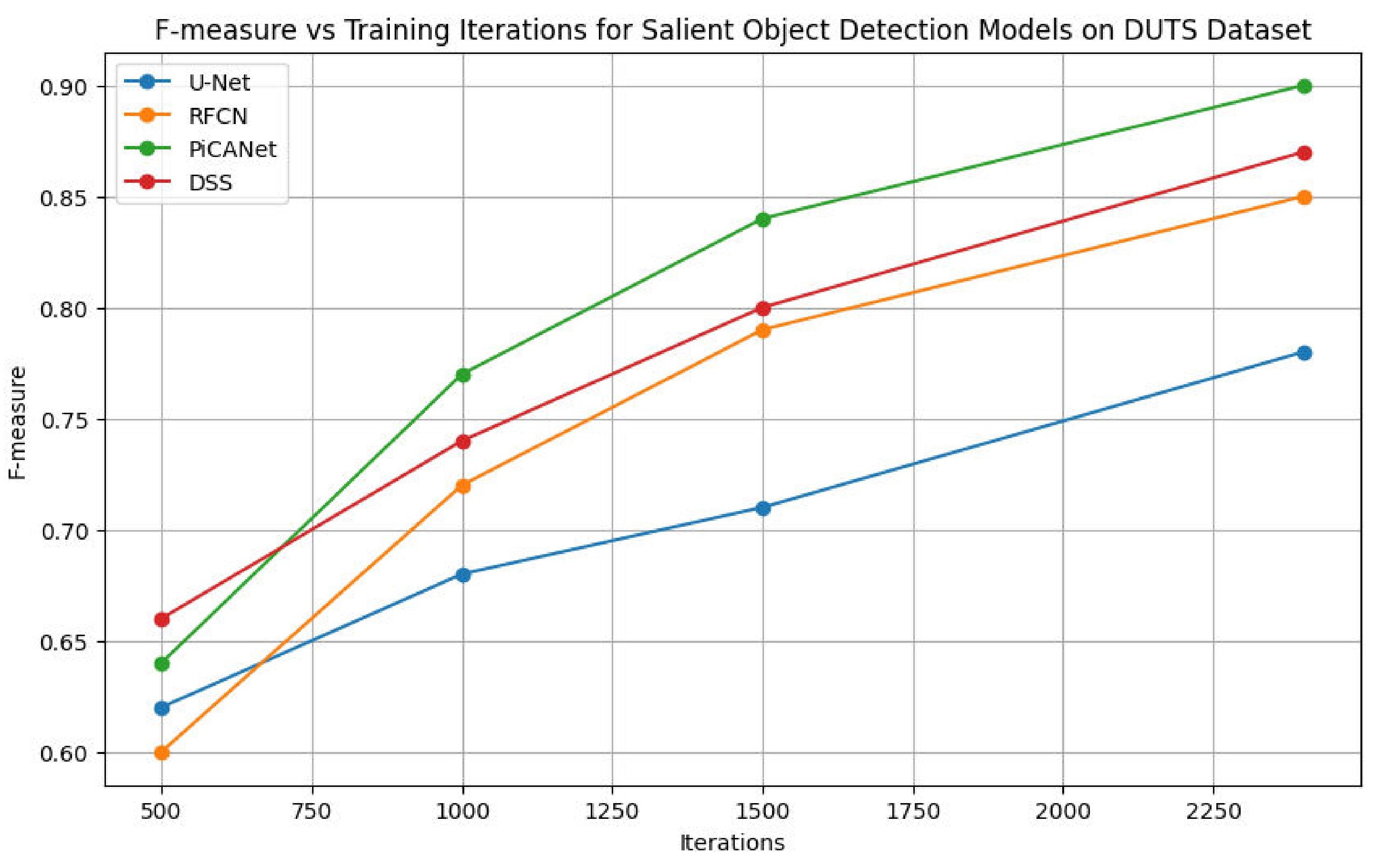}
  \caption{A comparative analysis of evaluation metrices for DUTS Dataset}
\end{figure}

\clearpage

\subsubsection{Autonomous Driving}
Autonomous driving combines computer vision and robotics to develop intelligent vehicles capable of driving without human intervention [147][148]. These systems rely on deep learning algorithms to process sensor data, including images, videos, LiDAR [149], and radar, for real-time localization, perception, planning, and control. Key tasks include object detection, lane detection, semantic segmentation, pedestrian detection, vehicle tracking, and trajectory prediction.

Object detection is critical for identifying pedestrians, bicycles, and vehicles. Advanced models like YOLO (You Only Look Once)[150] and SSD[151] (Single Shot Detector) achieve high-speed detection using encoder-decoder networks, while Faster R-CNN provides accurate results with region proposals. Transformer-based models like DETR enhance object detection by leveraging global relationships in images[152]. Semantic segmentation is also essential for understanding road scenes at a pixel level[153]. Early models like Fully Convolutional Networks (FCNs) were improved by DeepLabv3+ [154] and SegNet [155], while transformer-based SegFormer further advanced segmentation by incorporating attention mechanisms and multiscale feature extraction. Lane detection in autonomous driving relies on deep learning algorithms[156] such as SCNN, which captures spatial dependencies in lane structures. LaneNet uses instance segmentation to distinguish different lanes [157][158], while Ultra Fast Lane Detection (UFLD)[159] achieves real-time inference with high accuracy. Multi-task learning approaches like ENet-SAD integrate semantic segmentation for lane detection, improving overall efficiency.

Trajectory prediction is vital for anticipating the movements of dynamic objects like vehicles and pedestrians [160]. Recurrent Neural Networks (RNNs) and Long Short-Term Memory (LSTMs) are commonly used for sequence-based prediction. Advanced models like Graph Neural Networks (GNNs) consider agent interactions, while transformer-based models like TrajFormer use self-attention mechanisms to simultaneously learn spatial and temporal dependencies. Datasets are crucial for training and evaluating autonomous driving models. The KITTI dataset [161], widely used by researchers, provides stereo images, LiDAR point clouds, and GPS data for tasks like object detection, segmentation, and depth estimation. Model performance is assessed using various metrics: mean Average Precision (mAP) for object detection, Intersection over Union (IoU) for semantic segmentation accuracy, and Average Displacement Error (ADE) and Final Displacement Error (FDE) for trajectory prediction. Real-time performance is often evaluated based on frames per second (FPS).

\begin{table}[ht]
  \scriptsize
  \centering
  \begin{tabularx}{\textwidth}{|X|X|X|X|X|X|}
  \hline
  \textbf{Algorithm} & \textbf{Task} & \textbf{mAP} &\textbf{IoU(\%)} &\textbf{Inference Time (ms)} &\textbf{Model Size (MB)}\\ \hline
  \textbf{YOLOv4} & Object Detection & 91.4 & - & 22 & 240\\ \hline
  \textbf{Faster R-CNN} & Object Detection & 94.2 & - & 120 & 512\\ \hline
  \textbf{Mask R-CNN} & Object Detection+ Segmentation & 92.8 & - & 150 & 600\\ \hline
  \textbf{UNet} & Semantic Segmentation & - & 83.2 & 20 & 30\\ \hline
  \textbf{LaneNet} & Lane Detection & - & 89.1 & 18 & 50\\ \hline
  \end{tabularx}
  \caption{A comparative analysis of evaluation metrices for KITTI Dataset}
\end{table}

\subsubsection{Face Detection}
Face detection is a key problem in computer vision with applications in biometrics, surveillance, and human-computer interaction [162][163][164]. The objective is to accurately locate human faces in images or videos, a critical step in face recognition systems. Deep learning, particularly Convolutional Neural Networks (CNNs), has significantly advanced face detection accuracy and robustness under challenges like occlusion [165], varying lighting, and orientation [166]. Pioneering models like MTCNN [167] provide scalable solutions, while region-based detectors like Faster R-CNN have been adapted for high-precision face detection. For real-time applications, single-shot detectors such as YOLO [168] and SSD [169] are widely used. More recent models like RetinaFace enhance accuracy through context modules and landmark localization [170]. Additionally, transformer-based architectures like DETR [82] offer robust detection by modeling global image relationships. Lightweight networks like MobileNet-SSD support face detection on resource-constrained devices, enabling edge deployment. These advancements collectively drive progress in building efficient and accurate face detection systems across diverse environments.

The datasets in this regard significantly contribute to the training and evaluation of face detection algorithms. The WIDER FACE [171] dataset is also one of the most challenging benchmarks as it provides images with occluded, or very small and poorly lit faces. It has three sub-set sizes such as easy, medium, and hard to allow for evaluation criteria at different levels of difficulty.
Commonly, Precision, Recall, and F1-score metrics are used for the face detection evaluation to estimate the quality of the detection. The average precision (AP) and mean average precision (mAP) are the measures used for ranking the models. The models have also been evaluated for real-time performance measured in frames per second (fps).

\begin{table}[ht]
  \scriptsize
  \centering
  \begin{tabularx}{\textwidth}{|X|X|X|X|X|X|}
  \hline
  \textbf{Algorithm} & \textbf{Precision} & \textbf{Recall} &\textbf{F1-Score} &\textbf{mAP} &\textbf{FPS}\\ \hline
  \textbf{MTCNN} & 0.904 & 0.880 & 0.891 & 0.883 & 15\\ \hline
  \textbf{Faster R-CNN} & 0.936 & 0.885 & 0.910 & 0.891 & 8\\ \hline
  \textbf{YOLOv4} & 0.879 & 0.860 & 0.865 & 0.865 & 62\\ \hline
  \textbf{SSD} & 0.873 & 0.840 & 0.854 & 0.848 & 30\\ \hline
  \textbf{RetinaFace} & 0.940 & 0.910 & 0.925 & 0.905 & 18\\ \hline
  \textbf{DETR} & 0.913 & 0.900 & 0.907 & 0.890 & 12\\ \hline
  \textbf{MobileNet-SSD} & 0.860 & 0.840 & 0.850 & 0.850 & 34\\ \hline
  \end{tabularx}
  \caption{A comparative analysis of evaluation metrices for WIDER FACE Dataset}
\end{table}

\begin{figure}[ht]
  \centering
  \includegraphics[width=0.9\textwidth, height=3in]{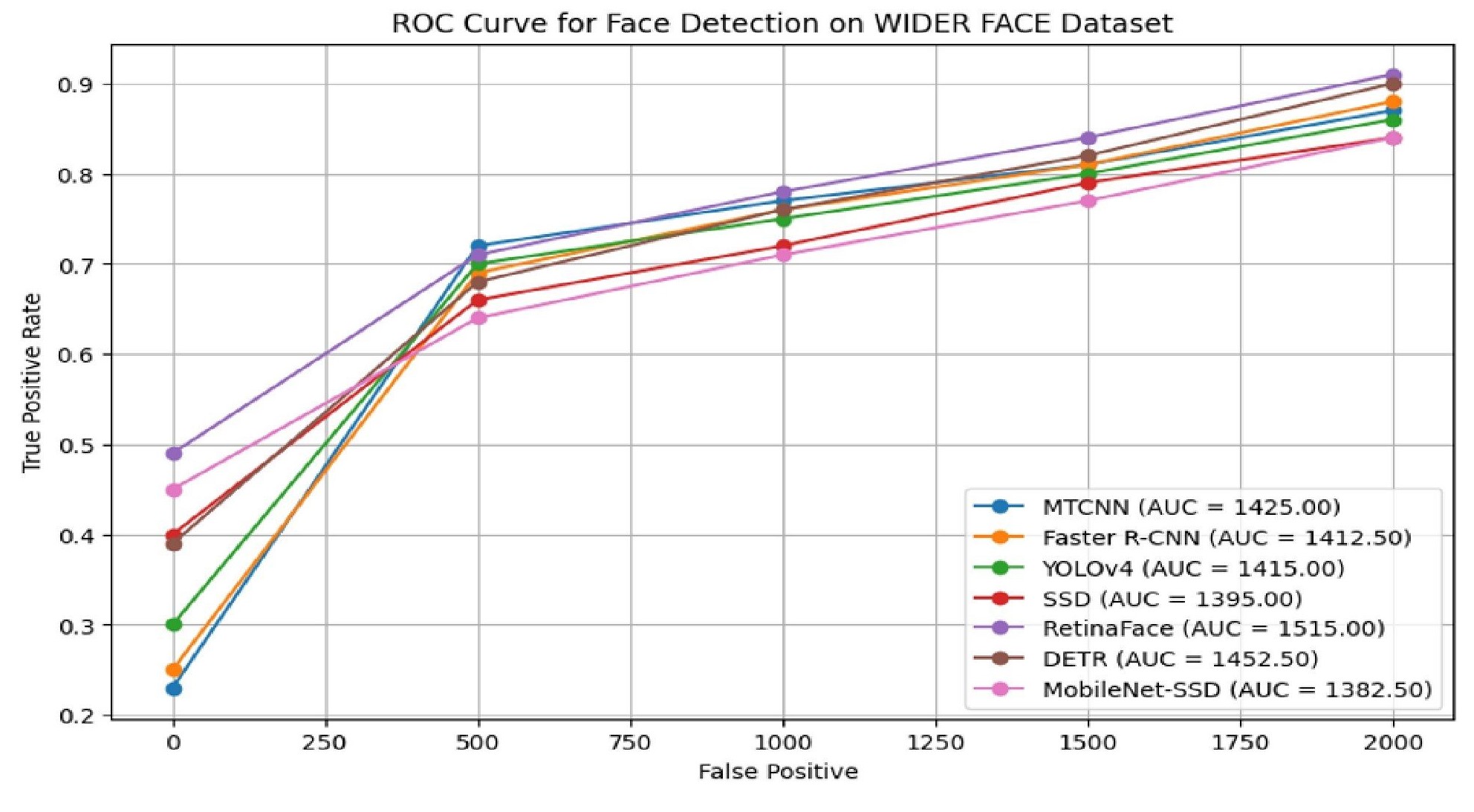}
  \caption{A comparative analysis of evaluation metrices for WIDER FACE Dataset}
\end{figure}

\subsubsection{Face recognition}
Face recognition involves identifying individuals by analyzing facial features and is widely used in biometrics, data security, and social networks [172][173][174]. Advances in deep learning, particularly Convolutional Neural Networks (CNNs) and transformer-based models, have significantly improved face recognition accuracy. Early systems like DeepFace used deep CNNs to extract key facial components [175]. FaceNet [176] introduced triplet-loss learning to minimize distances between facial representations. VGGFace adopted deeper network structures to improve feature extraction and recognition precision [177]. DeepID4 advanced identity-discriminative features by employing multiple CNNs for richer facial attribute representation [178]. SphereFace introduced angular margin loss functions to enhance intra-class compactness and inter-class separability [179]. CosFace and ArcFace further refined this by using cosine and additive angular margins, respectively, achieving strong performance even under challenging conditions [180][181]. These developments have collectively propelled face recognition into a highly accurate and robust technology suitable for real-world deployment.

In more recent developments Vision Transformers (ViT) and Swin Transformers that are based on the transformer architectures, have also proven to be effective in retrieving long range dependencies and the global context [182][183]. Such models have also been applied to CNN-based models as a hybrid approach to local and global feature extraction enhancement.

\begin{table}[ht]
  \scriptsize
  \centering
  \begin{tabularx}{\textwidth}{|X|X|X|X|}
  \hline
  \textbf{Algorithm} & \textbf{Accuracy(↑)} & \textbf{TAR@FAR=0.1\%(↑)} &\textbf{EER(↓)}\\ \hline
  \textbf{DeepFace} & 97.35\% & 91.50\% & 3.5\%\\ \hline
  \textbf{FaceNet} & 99.63\% & 97.00\% & 0.9\%\\ \hline
  \textbf{VGGFace} & 98.95\% & 95.20\% & 1.2\%\\ \hline
  \textbf{DeepID} & 98.97\% & 95.40\% & 1.1\%\\ \hline
  \textbf{SphereFace} & 99.42\% & 96.90\% & 0.8\%\\ \hline
  \textbf{CosFace} & 99.73\% & 97.40\% & 0.7\%\\ \hline
  \textbf{ArcFace} & 99.83\% & 98.10\% & 0.6\%\\ \hline
  \textbf{ViT} & 99.70\% & 97.80\% & 0.7\%\\ \hline
  \end{tabularx}
  \caption{A comparative analysis of evaluation metrices for Labeled Faces in the Wild (LFW) Dataset}
\end{table}

\begin{figure}[ht]
  \centering
  \includegraphics[width=0.9\textwidth, height=3in]{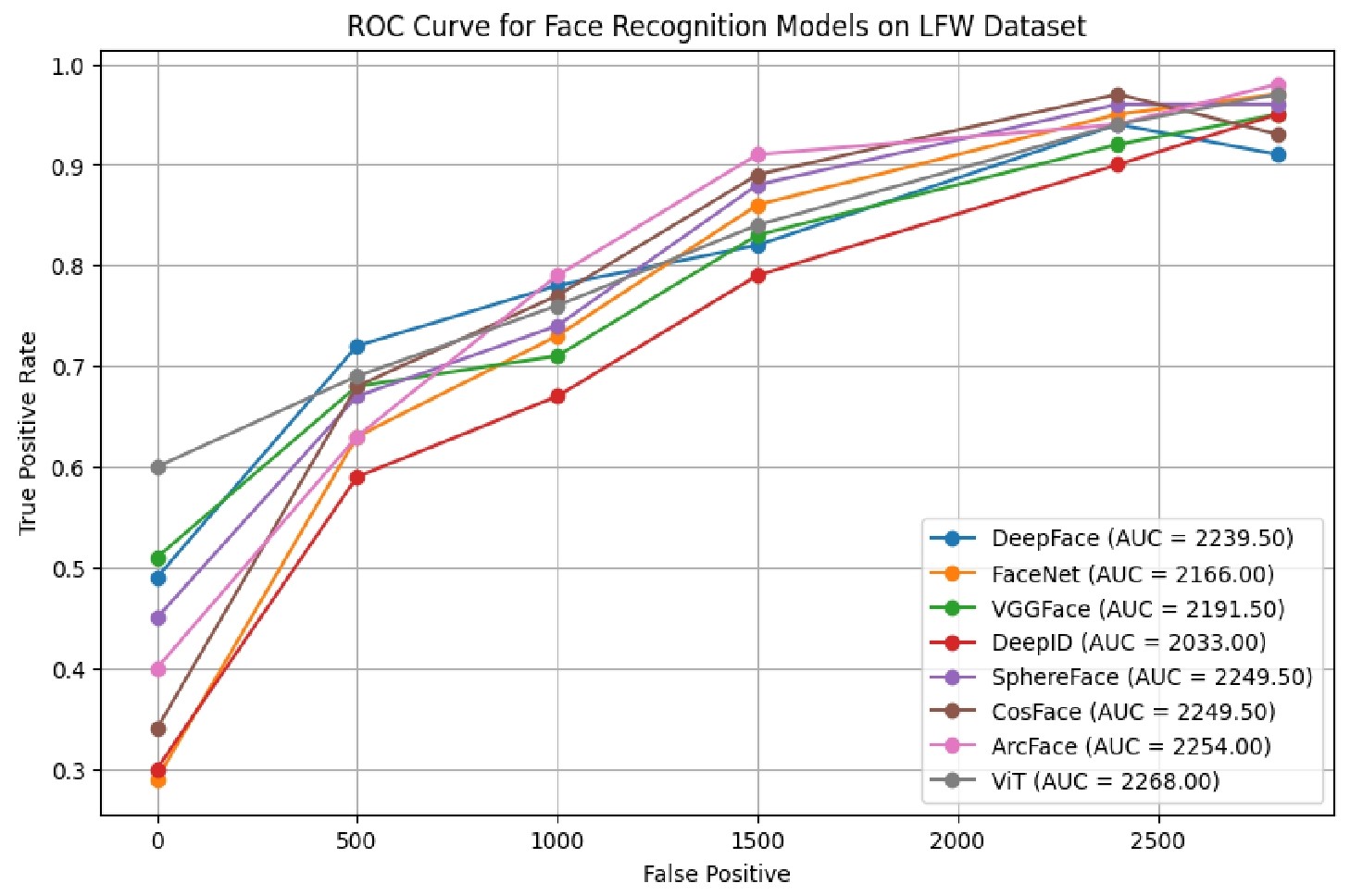}
  \caption{A comparative analysis of evaluation metrices for DUTS Dataset}
\end{figure}

The Labeled Faces in the Wild (LFW) Dataset [184] serves as a yardstick for the performance of various face recognition algorithms. It has a total of 13,233 images of 5749 subjects, and has been created to allow for the validation of face recognition work in the context of uncontrolled settings, that is, when there is a change in pose, lighting, occlusions and other factors.
Face recognition models are estimated in terms of Accuracy, TAR and FAR. Furthermore, EER measure the extent to which TAR and FAR affect each other in frequency of reporting.

\section{Future Scope}
Although real-time object detection has progressed rapidly, several challenges remain unresolved, and many opportunities for advancement are yet to be explored before these systems can achieve widespread deployment in real-world scenarios. This section highlights key research directions that warrant further investigation.

First, there is an urgent need for standardized benchmark architectures that simultaneously account for accuracy, latency, and energy consumption across heterogeneous hardware platforms such as GPUs, NPUs, and edge TPUs, thereby ensuring fair comparisons and reproducibility. Second, despite the continuous evolution of YOLO and related variants, real-time detectors still struggle with small, low-contrast, and occluded objects. Future work should therefore emphasize improved feature fusion, attention-driven modules, and multimodal strategies to enhance robustness under these challenging conditions. Third, while transformer-based detectors (e.g., DETR and its successors) introduce a conceptually elegant framework, their computational cost limits real-time deployment; thus, the design of lightweight and hardware-aware transformer models remains a promising avenue.

Moreover, the integration of detection with temporal reasoning and object tracking under strict real-time constraints offers another fertile direction for research. Finally, in safety-critical domains such as healthcare, autonomous driving, and industrial inspection, there is an increasing demand for domain adaptation, interpretability, and reliability to ensure trustworthy deployment. Beyond these areas, numerous additional challenges related to scalability, optimization, and real-world adaptability continue to present opportunities for further exploration.

\section{Conclusion}
Real-time object detection using deep learning has a broad spectrum of applications, including autonomous driving, surveillance, healthcare, robotics, smart agriculture, and smart cities. This survey presents a comprehensive analysis of various deep neural network architectures, each with distinct strengths and limitations. Notable advancements, such as YOLO, SSD, and transformer-based models, have contributed significantly to improving both detection accuracy and processing speed. The article systematically reviews state-of-the-art object detection frameworks and highlights recent developments in deep learning techniques. It also compiles a diverse set of benchmark datasets, providing a solid foundation for evaluating model performance. In addition, the survey explores a wide range of real-world applications and includes a comparative analysis of different deep learning models. Through this investigation, valuable insights are gained that inform the design and optimization of neural networks for object detection. The study identifies key challenges and trends, fostering innovative ideas for future system development. Moreover, it outlines strategic directions for further research, contributing to a deeper understanding of the object detection landscape. The findings of this work offer meaningful guidance for both academic researchers and industry practitioners, emphasizing areas for improvement and encouraging continued progress in the field of real-time object detection using deep learning.

\clearpage

\bibliographystyle{unsrt}  


\end{document}